\documentclass[letterpaper]{article} % DO NOT CHANGE THIS
\usepackage{aaai25}  % DO NOT CHANGE THIS
\usepackage{times}  % DO NOT CHANGE THIS
\usepackage{helvet}  % DO NOT CHANGE THIS
\usepackage{courier}  % DO NOT CHANGE THIS
\usepackage[hyphens]{url}  % DO NOT CHANGE THIS
\usepackage{graphicx} % DO NOT CHANGE THIS
\usepackage{natbib}  % DO NOT CHANGE THIS AND DO NOT ADD ANY OPTIONS TO IT
\usepackage{caption} % DO NOT CHANGE THIS AND DO NOT ADD ANY OPTIONS TO IT
\usepackage{algorithm}
\usepackage{algorithmic}

\usepackage{newfloat}
\usepackage{listings}
\DeclareCaptionStyle{ruled}{labelfont=normalfont,labelsep=colon,strut=off} % DO NOT CHANGE THIS
\floatstyle{ruled}
\newfloat{listing}{tb}{lst}{}
\floatname{listing}{Listing}
\usepackage{amssymb}
\usepackage{adjustbox}
\usepackage{mathtools}
\usepackage{multirow}
\usepackage{xcolor}
\usepackage{subfig}
\usepackage{enumitem}
\usepackage{subcaption}
\usepackage{booktabs}
\newtheorem{theorem}{Theorem}[section]

\newtheorem{definition}[theorem]{Definition}

\title{Incorporating Human Flexibility through Reward Preferences \\ in Human-AI Teaming}
\author{
    Siddhant Bhambri\textsuperscript{\rm 1}\equalcontrib,
    Mudit Verma\textsuperscript{\rm 1}\equalcontrib,
    Upasana Biswas\textsuperscript{\rm 1},
    Anil Murthy\textsuperscript{\rm 1},
    Subbarao Kambhampati\textsuperscript{\rm 1}
}
\affiliations{
    \{siddhantbhambri, muditverma, ubiswas2, abmurthy, rao\}@asu.edu \\
    \textsuperscript{\rm 1}School of Computing and AI\\
    Arizona State University\\
    Tempe, AZ, USA\\
}

\usepackage{bibentry}
\begin{document}

\maketitle

\begin{abstract}
Preference-based Reinforcement Learning (PbRL) has made significant strides in single-agent settings, but has not been studied for multi-agent frameworks. On the other hand, modeling cooperation between multiple agents, specifically, Human-AI Teaming settings while ensuring successful task completion is a challenging problem. To this end, we perform the first investigation of multi-agent PbRL by extending single-agent PbRL to the two-agent teaming settings and formulate it as a Human-AI PbRL Cooperation Game, where the RL agent queries the human-in-the-loop to elicit task objective and human's preferences on the joint team behavior. Under this game formulation, we first introduce the notion of Human Flexibility to evaluate team performance based on if humans prefer to follow a fixed policy or adapt to the RL agent on the fly. Secondly, we study the RL agent's varying access to the human policy. We highlight a special case along these two dimensions, which we call Specified Orchestration, where the human is least flexible and agent has complete access to human policy. We motivate the need for taking Human Flexibility into account and the usefulness of Specified Orchestration through a gamified user study. We evaluate state-of-the-art PbRL algorithms for Human-AI cooperative setups through robot locomotion based domains that explicitly require forced cooperation. Our findings highlight the challenges associated with PbRL by varying Human Flexibility and agent's access to the human policy. Finally, we draw insights from our user study and empirical results, and conclude that Specified Orchestration can be seen as an upper bound PbRL performance for future research in Human-AI teaming scenarios.
\end{abstract}


\section{Introduction}
\label{sec:introduction}

\begin{figure}
    \centering
    \includegraphics[width=0.8\linewidth]{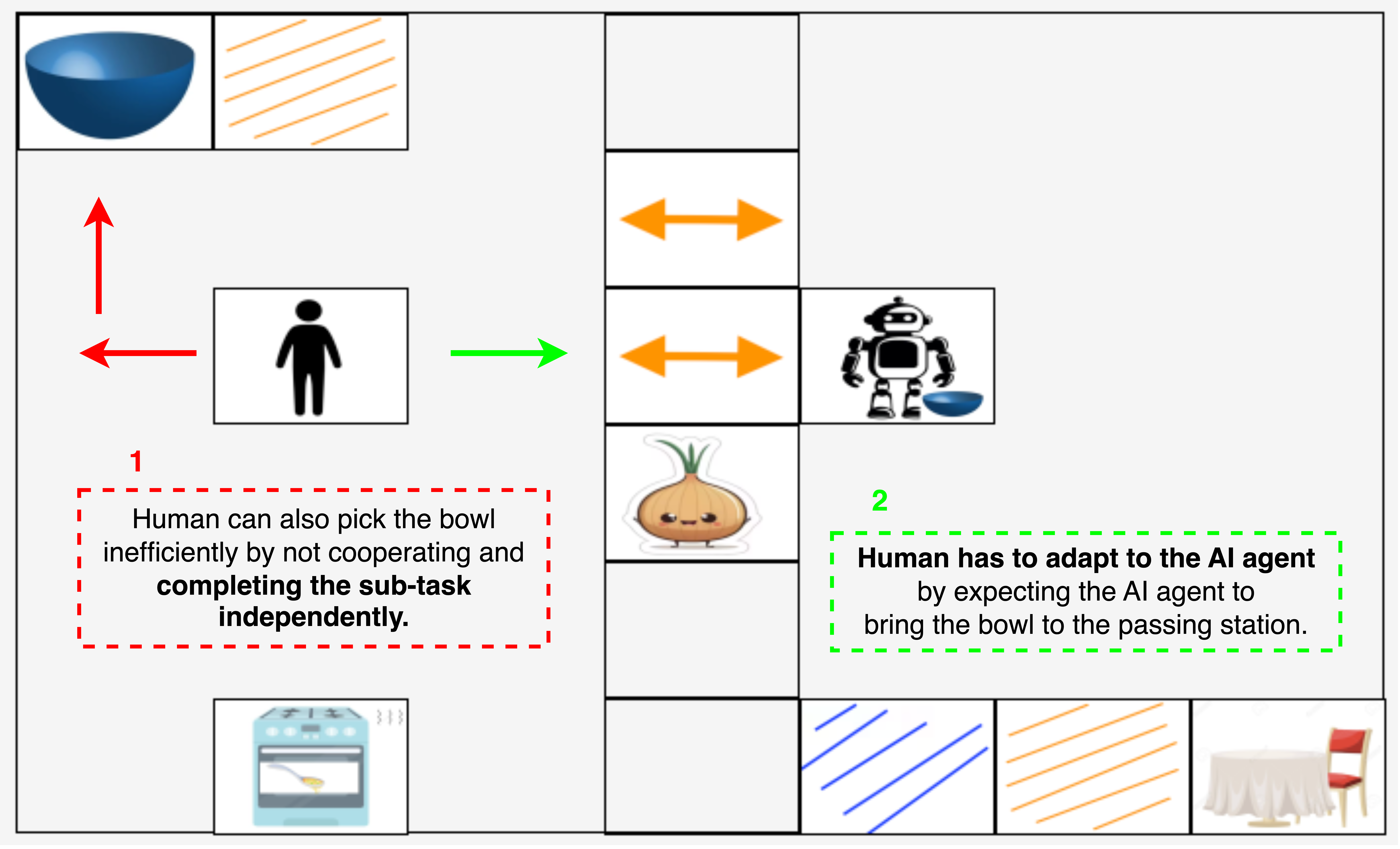}
    \caption{In this example task, a human and an AI agent need to cook a soup and deliver to the table. Note, that human can either \textcolor{green}{continuously adapt to the AI agent} by keeping an expectation on its actions to complete the sub-task or plating the soup in a bowl efficiently, or if possible, \textcolor{red}{complete it independent of the AI agent} which is less cognitively demanding but inefficient. We aim to bridge this gap by investigating Human-Flexibility in Human-AI Teaming and showcase how PbRL can be useful.}
    \label{fig:study:grid}
\end{figure}

Preference-based Reinforcement Learning (PbRL) has been studied in several complex domains like robotic manipulation and locomotion to incorporate human preferences on AI agent's behavior \cite{wirthsurvey}. The initial conceptualization of PbRL has primarily focused on single-agent dynamics with a human-in-the-loop observer available to provide feedback on agent queries \cite{christiano2017deep}. While several methods have explored improving reward learning and reducing feedback queries, investigating PbRL for multi-agent scenarios remains uncharted.

Modeling cooperation for Human-AI Teaming scenarios has become all the more relevant with the increase in the pivotal roles AI agents play in day-to-day human activities \cite{wachter2017transparent,maedche2019ai}. Typically, cooperative teaming requires all acting agents to understand each other's strategies (via explicit communication or implicit deduction from past behavior) and execute compatible behavior boosting team performance (see Figure \ref{fig:study:grid}). While multi-agent cooperation is hard \cite{du2023cooperative}, the problem is even more challenging when one of the agents is a human \cite{gal2022multi,carroll2019utility}. It becomes crucial for the AI agent to consider human's cognitive limitations, their preferences on the robot's behavior, and align its policies to be compatible with policies that humans deem achievable. However, as we emphasize later, most of the current approaches to modeling Human-AI teaming leave the burden of adapting on the human \cite{carroll2019utility,fcp,hsp}, which often leads to increased cognitive load.

Specifically, in the case of Human-AI teaming, it is imperative to discuss, both, 1) the human-in-the-loop's expectations or preferences on the joint team behavior, and 2) their constraints or assumptions on their own policy (or policies). To this end, we provide a first exploration of a more general perspective of PbRL extending to the two-agent, i.e., the human-AI team setting where the human-in-the-loop co-exists in the same environment as the AI agent and jointly performs actions for achieving the team goal. We formulate this problem as a Human-AI PbRL Cooperation Game, where the human's feedback can act as a mechanism for the AI agent to elicit task objectives and human's preferences on the team behavior. 

One might argue that there may exist many solution policy pairs to the Human-AI PbRL Cooperation Game, in which the two agents can cooperate with each other leading to successful task completion. However, for a human-AI team setting (such as in Figure \ref{fig:study:grid}), it is commonly assumed that training an AI agent to learn a \textit{universal best-response} policy leads to optimal team performance, which imperatively leads to the human bearing the burden of adaptation \cite{carroll2019utility,fcp,hsp} (for example, the \textcolor{green}{green path} shown in Figure \ref{fig:study:grid}). A key aspect of modeling this adaptation from the human's perspective is to study the spectrum of flexibility to answer if the human partner is willing to execute a fixed policy or is willing to adapt. Hence, we introduce the notion of Human Flexibility which is defined as a subset of the solution set for the Human-AI PbRL Cooperation Game, consisting of policies that the human is willing to execute for the team task, and we motivate this via a gamified user study conducted over 20 lay users on the task shown in Figure \ref{fig:study:grid}, drawing a systematic comparison between the two cases. We find that 50\% of the participants complete the cooperative task by not choosing to adapt to the AI agent, while 100\% task success is seen when they follow a pre-determined fixed policy. On average, participants took significantly more time to execute their next action while adapting to the other agent. We also noted higher levels of frustration and cognitive load when participants did not follow a fixed policy.

Operating under this setting, we also argue along the aspect of Access to Human Policy, which is costly to obtain, but required for learning the reward function and the corresponding optimal agent policy. If the agent has zero access to the human's policy (as selected by the human agent to execute), it has an additional challenge of `imagining' human's actions when it queries the human for their preference on the team behavior. On the other hand, the problem can be simplified to a single-agent PbRL setting if the agent has complete knowledge of the human policy and the human has a single policy in their set of feasible team strategies. We call this case as Specified Orchestration, which requires maximal information regarding human behavior and arguably the simplest case for the AI agent, therefore, can be treated as a loose upper bound to the performance of Human-AI PbRL algorithms. Finally, we re-purpose a suite of domains, comprising of variants of the Highway domain \cite{highway-env} and adapted MuJoCo locomotion domains \cite{openaigym}, that require forced cooperation in a two-agent setting with one agent assumed to be a human agent in our case. We adapt existing state-of-the-art single agent PbRL methods (i.e. RUNE \cite{rune}, SURF \cite{surf} and PEBBLE \cite{pebble}) and extend them to this two-agent setting and draw insights that we believe can be useful for future research in Human-AI teaming scenarios. We summarize the contributions of this work as follows:
\begin{enumerate}
    \item As a first exploration of PbRL beyond a single-agent setting, we investigate the multi-agent Human-AI teaming setting, and formulate this problem as a Human-AI PbRL Cooperation Game. 
    \item We introduce the notion of Human Flexibility and conduct a human subject study showcasing how lay users find following a pre-determined rigid policy (less flexible case) much easier than understanding the agent's behavior and adapting on the fly (more flexible case). We also discuss the possible cases by varying the agent's Access to Human Policy.
    \item We extend state-of-the-art PbRL algorithms for our setting, compare their performance over a suite of two-agent domains, and draw insights on the challenges associated with utilizing PbRL for Human-AI cooperation.
\end{enumerate}

\section{Related Work}
\label{sec:related_work}

% \label{subsec: related_pbrl}
\textbf{Preference-based Reinforcement Learning:} PbRL has become a popular paradigm for human-in-the-loop settings where the human is queried for binary feedback by the AI agent to learn a reward function \cite{furnkranz2012preference,christiano2017deep,ibarz2018reward,bewley2021interpretable,bpref}. These methods have been effective in teaching an AI agent to accomplish a task where rewards may be otherwise hard to specify, and also to incorporate the human-in-the-loop's preferences on how the task needs to be achieved \cite{christiano2017deep}. However, PbRL methods have only been seen in single-agent settings where the AI agent is acting in the environment, and the human is only interacting with the system as and when asked for query feedback \cite{pebble,surf,rune}. Having realized the potential of PbRL frameworks, we see its applicability in human-AI teaming scenarios where both the human and the AI agent act in the environment to accomplish a team task, and the human can specify their preferences through binary feedback distilled into the agent's reward function. Closest to our research are prior works in multi-agent Inverse RL \cite{natarajan2010multi,adams2022survey,pynadath2002communicative}. We note that IRL and PbRL have key distinctions in assuming expertise of the human-in-the-loop and feedback modality. Several works have attempted to study multi-agent IRL for homogeneous agent systems \cite{bogert2017scaling,lin2019multi,lin2017multiagent} whereas our focus is on a heterogeneous system with a two-player Human-AI cooperative team. \cite{cirl} proposed the Cooperative IRL formulation to highlight the limitations of classic formulation which does not allow humans to provide teaching demonstrations. In contrast, we focus on allowing the AI agent to make explicit queries to the human agent to elicit preferred behaviors. While IRL-based setups are dependent on expert demonstrations by the human (which may not always be feasible), PbRL, on the other hand, makes it easier to provide a binary feedback over agent trajectories.

\textbf{Multi-Agent Teaming :} Deep Reinforcement Learning has also been used in multi-agent settings \cite{nguyen2020deep,du2023cooperative} such as non-competitive games \cite{lowe2017multi,gupta2017cooperative}, zero-sum \cite{lin2017multiagent}, general-sum games \cite{lin2019multi,lerer2019learning} and other collaborative environments where rewards are not shared across agents \cite{leibo2017multi}. Several works have investigated the challenges in Human-AI teams but commonly assume robot's access to the team reward function \cite{carroll2019utility,fcp,tylkin2021learning}. Prior works have also studied joint-decision making in a two-agent collaboration setting \cite{crandall2018cooperating,dimitrakakis2017multi,radanovic2019learning,DBLP:journals/corr/NikolaidisNPS17} but do not consider inverse RL or PbRL setting. 

% \textcolor{red}{need to add MA IRL works}

\section{Preliminaries}
\label{sec:preliminaries}

\label{subsec: prelim_pbrl}
\textbf{Preference-based Reinforcement Learning : } In traditional PbRL setups, an agent $A$ interacts with the environment $\mathcal{E}$ with a missing reward function $\mathcal{R}$. It assumes the underlying problem as a MDP $\mathcal{M} = \langle \mathcal{S}, \mathcal{A}, \mathcal{T}, \mathcal{R}, \gamma \rangle$. The agent $A$ may not have access to this reward system $\mathcal{R}$ as it may either be hidden \cite{ng2000algorithms}, hard to specify as seen in tacit-control tasks \cite{christiano2017deep}, or otherwise may not be able to capture the goal completely as seen in human-in-the-loop scenarios \cite{goodrich2008human}.

The problem of learning the unknown reward function $r_\psi$, in single-agent settings, is formulated as a supervised learning problem \cite{christiano2017deep}. A classifier is trained on limited human feedback over agent trajectories. Formally, each agent trajectory is a segment $\sigma$, a sequence of agent observations and actions $\{(s_k, a_k), (s_{k+1}, a_{k+1}), \cdot \cdot \cdot (s_{k+N-1}, a_{k+N-1})\}$ where $N$ represents the horizon. The human-in-the-loop, i.e., the teacher, gives a feedback $y \in \{0, 1\}$ to the agent when queried with a pair of trajectories ($\sigma_1, \sigma_2$), and this feedback is stored as a label along with the two segments in a trajectory buffer $\mathcal{D}$ as ($\sigma_1, \sigma_2, y$). Following the Bradley-Terry preference model \cite{bradley1952rank}, the preference predictor using the reward function $r_\psi$ is learned:
\begin{equation}
 \label{eq:prefprediction}
     P_{\psi}[\sigma_0 \succ \sigma_1] = \frac{\exp\Bigl(\sum_{t}\hat{r_\psi}(s_t^0, a_t^0)\Bigl)}{\sum_{i \in \{0,1\}} \exp\Bigl(\sum_{t}\hat{r_\psi}(s_t^i, a_t^i)\Bigl)}
 \end{equation}

where, $[\sigma_0 \succ \sigma_1]$ implies that trajectory segment $\sigma_0$ is preferred over trajectory segment $\sigma_1$ by the human teacher. Given the trajectory buffer $\mathcal{D}$, the reward function $r_\psi$ is learned by training on minimizing the following Cross-Entropy loss ($\mathcal{L}_{CE}$):
\begin{equation}
\label{eq:cross_entropy}
\begin{split}
     -\displaystyle \mathop{\mathbb{E}}_{\substack{(\sigma_0, \sigma_1, \\ y) \sim \mathcal{D}}} \Bigr[y(0)\text{log}P_\psi[\sigma_0 \succ \sigma_1] + y(1)\text{log}P_\psi[\sigma_1 \succ \sigma_0]\Bigr]
\end{split}
\end{equation}

\section{PbRL for Human-AI Teaming}
\label{sec:pbrl_teaming}

% \textcolor{red}{prelims discuss pbrl / and coopeartion game is POMDP with missing agent rewards.}

% Preference-based RL has been predominantly studied in single-actor human-observer settings where the human provides feedback on the agent's behavior but does not act in the shared environment. The aim of this work is to explore PbRL for Human-AI Teaming. We consider a setup where the human and the AI agents are a cooperative team and the role of the robot is to achieve goals as set by the human agent. Having no knowledge of its objectives, the robot relies on a reward learning process through PbRL to elicit and learn about the task objective and human preferences over the team behavior. We formalize our Human-AI Team PbRL as a special case of the CIRL formulation \cite{cirl} (Section \ref{subsec:prelim_markov_game}). Further, we discuss the two critical aspects for Human-AI Teaming i.e. Human Flexibility (Section \ref{subsec:human_flexibility}), assumption on agent's access to human policy (Section \ref{subsec:policy_access}), and a special case of Specified Orchestration (Section \ref{subsec:spec_coop}).

Preference-based RL has been predominantly studied in single-actor human-observer settings where the human provides feedback on the agent's behavior but does not act in the shared environment. The aim of this work is to explore PbRL for Human-AI Teaming. We consider a setup where the human and the AI agents are a cooperative team and the role of the robot is to achieve goals as set by the human agent. Having no knowledge of its objectives, the robot relies on a reward learning process through PbRL to elicit and learn about the task objective and human preferences over the team behavior. In this section, we formalize our Human-AI Team PbRL as a special case of the CIRL formulation \cite{cirl}. Further, we discuss the two critical aspects of Human-AI Teaming, i.e., Human Flexibility - the assumption on the agent's access to human policy, and a special case of Specified Orchestration.

\subsection{Human-AI PbRL Cooperation Game}
\label{subsec:prelim_markov_game}

\cite{cirl} introduces the problem of Cooperative Inverse Reinforcement Learning but studies it for Apprenticeship Learning. For our Human-AI Team PbRL setting, we modify their formulation as follows: 

\begin{definition}
    A Human-AI PbRL Cooperation Game, $M$ is a two-player Markov game with identical payoffs between a human $H$ and a robot $R$. The game is defined as a tuple $M = (S, \{{\mathcal{A}^H, \mathcal{A}^R} \}, T(\cdot | \cdot, \cdot, \cdot), R_\theta(\cdot,\cdot,\cdot), \gamma, \mathcal{H}_\Pi, \mathcal{H}_\mathcal{B})$ with the following definitions: 
    \begin{itemize}
        \item $S$ a set of world states: $s \in S$;
        \item $\mathcal{A}^H$ a set of actions for H : $a^H \in \mathcal{A}^H$;
        \item  $\mathcal{A}^R$ a set of actions for R : $a^R \in \mathcal{A}^R$;
        \item $T(\cdot | \cdot, \cdot, \cdot)$ a conditional distribution on the next world state, given previous state and action for both agents, i.e., $T(s' | s, a^H, a^R)$;
        \item $R_\theta(\cdot,\cdot,\cdot)$ a reward function that maps world states, joint actions to reward values according to H's preference on the team objective, H's objectives, R's objectives and team behavior;
        \item $\gamma$ a discount factor : $\gamma \in [0,1]$;
        \item $\mathcal{H}_\Pi$ Human Flexibility: $|\Pi_H|$, i.e., set of policies admissible to the human (see Definition \ref{def:h_flexible})
        \item $\mathcal{H}_\mathcal{B}$ Access to Human Policy: $\mathcal{H}_\mathcal{B} \in [0,1]$, i.e., R's budget for querying the human policy (see Definition \ref{def:h_access})
    \end{itemize}
\end{definition}

The reward function $R$ is set by the human agent defining the team objectives and human-agent acceptable policy pairs $(\pi_H, \pi_R)$. Note that in teaming settings, $R$ can be decomposed further into feasibility or environment constraints, task reward, and human preferences. However, due to their inherent limitation, PbRL frameworks have only looked at jointly learning a single reward function, and it is also non-trivial to combine multiple rewards otherwise. Now, consider that the game begins at the initial state $s_o \in S$. The human agent pre-computes acceptable solutions to the game with assumed knowledge on AI partner's capabilities. Suppose the computed solution set is $\Pi = (\pi_H, \pi_R)_0, (\pi_H, \pi_R)_1 \cdots (\pi_H, \pi_R)_n$. For simplicity, suppose the solution set is singleton, i.e. $|\pi_H|=1$, and the agent has complete access to human policy, i.e., $\mathcal{H}_\mathcal{B}=1$. The human takes an action following their policy $\pi_H$. The robot takes a random action in $\mathcal{A}^R$. As the game proceeds, robot builds a trajectory bank from past experiences. Following PbRL setup, the agent samples two trajectories from this bank and queries for human feedback. The human agent uses their internal reward function $R$ to provide their preference label. A behavior pair where the shown human actions align with the true human actions from $\pi_H$ and the shown robot actions align with human expected $\pi_R$ such that resultant return of the trajectory is higher than the alternate is preferred. While policies can be arbitrary functions of history to next action as $\pi_H : [\mathcal{A}^H \times \mathcal{A}^R \times S]* \rightarrow \mathcal{A}^H$ (and similarly for robot R), we restrict the policy to be defined on states (thereby assuming the dynamics, reward function and the optimal policy pair is Markovian w.r.t. the world state).

\subsection{Flexibility of the Human-in-the-Loop}
\label{subsec:human_flexibility}

It is important to note that for our human-AI teaming PbRL setup, we only consider the case where the AI agent's access to the underlying reward function is restricted. That is, the agent does not experience a sampled reward from the environment, and instead, depends on an explicit reward learning procedure through human feedback. Assume the set $\Pi = \{\Pi_1, \Pi_2, \cdots \Pi_n\}$ where $\Pi_i = (\pi_H, \pi_R)_i$ be the possible solutions of the Cooperation Game defined in Section \ref{subsec:prelim_markov_game} as can be computed by the human. We define Human Flexibility over their policy, i.e. $\mathcal{H}_\Pi$ as : 

\begin{definition}
    \label{def:h_flexible}
    Given the complete solution set $\Pi$ , the human-in-the-loop is said to be $\epsilon$-H-flexible if 
    \[
        \mathcal{H}_\Pi = \left|\{ {\pi_H}_k \, | \, k = 1 \, \text{to} \, n \}\right| = \epsilon - 1
    \]
\end{definition}

The definition of H-flexible (\ref{def:h_flexible}) human in the loop helps us recognize the rigidity of the human agent over a strategy for the two agents. A large $\epsilon \rightarrow |\Pi_H|$ implies that the human is willing to execute any one of the policies in $\Pi_H$ subsequently providing the robot to converge to a larger solution space (policies ${\pi_R}_k$). The nature and requirements of the cooperation for the given team-goal can be helpful in determining the values of $\epsilon$. For instance, $\epsilon = 0$ is useful when considering human cognitive limitations, thereby the human is able to come up with a single policy for themselves.

When $\epsilon > 1$, the human agent can select which policy ${\pi_H}_k$ they will follow. In this work, we assume that the AI agent's objective is to be conformant to all human policies. This choice is motivated from how prior works in Human-AI cooperation have attempted to learn the robot's policy \cite{fcp}. Other alternatives are: the agent is compatible to at least one of the human policies, a subset of the human policies etc. exist which would also impact the training regime of the RL agent (i.e. which policy is used by the human agent when the robot is learning its policy).

\subsection{Access to Human Policy}
\label{subsec:policy_access}
Based on the Human-AI Team PbRL Cooperation Game describe in Section \ref{subsec:prelim_markov_game}, the human and the robot take actions in a shared environment. This implies that for the duration of training, in addition to the preference feedback, the framework assumes the human to sample their optimal action $a_h$ for a world state $s$. That is, the agent gains indirect access to the human policy by virtue of the shared world state and the fact that both agents coexist in the shared world. While this is the default operating assumption through this work, it has higher requirements from the human agent as compared to single-agent PbRL. For example, in single-agent PbRL the human is required to be present in the loop during agent's training procedure and provide feedback, whereas in the current setup, in addition to the preference feedback, the human is also required to follow their policy throughout the training process (which can be millions of steps).

We can re-interpret the Human-AI Team PbRL Cooperation Game as follows: the agent has access to an environment simulator which can simulate the agent's actions and the human's actions. Now, the agent uses the policy $\pi_R$ being learned and an imagined human policy $\pi_R^h$ to generate team behaviors. These behaviors can be stored in a trajectory buffer and queried to the human actor. The advantage of this re-interpretation is that while the human agent is present in the loop, they are not required to execute millions of actions during agent's training. The disadvantage is, the agent has to model the human's policy in addition to the human's preferences. This re-interpretation of the game can enable us to study the performance of the PbRL agent when it has budgeted access to human actions.  Formally,
\begin{definition}
    \label{def:h_access}
    Given, the agent's total training budget of $N$ episodes, and the budget to query human policy for $\mathcal{Q}$ episodes, agent's access to human policy, $\mathcal{H}_\mathcal{B}$, is defined as:
    \[
        \mathcal{H}_\mathcal{B} = \mathcal{Q}\big/N,
    \]
\end{definition}
Full access is equivalent to the default interpretation, 50\% access is when the agent can query for the human's actions for the first half of the training process and 0\% access implies the agent cannot query human's actions at all. In this work, the agent defaults to using random actions as imagined human actions when it exhausts the human-action query budget. Techniques like Behavior Cloning \cite{carroll2019utility} can also be leveraged to independently model the human's policy along with the PbRL step in future work.

\subsection{Specified Orchestration}
\label{subsec:spec_coop}

It is clear that since the human agent defines the team objective through reward function $R$, they are restricting the possible robot behaviors to achieve the team goal. It becomes extremely challenging to learn this reward function if the agent does not have access to the human's policy. Further, as described in Section \ref{subsec:human_flexibility}, if the human is switching their policy in the feasible team behavior set during the agent's training process, the agent now has an additional objective of learning a conformant response policy that is compatible with all of the policies in $\epsilon$-Human Flexible solution set. Empirical results in Section \ref{sec:expts_results} show that the agent's performance in policy learning is significantly diminished in the absence of these assumptions. Hence, we formally call this constrained case Specified Orchestration, and it can be defined as:
\begin{definition}
    Specified Orchestration is a special case of the Human-AI Team Cooperation Game $M = (S, \{{\mathcal{A}^H, \mathcal{A}^R} \}, T, R_\theta, \gamma, \mathcal{H}_\Pi, \mathcal{H}_\mathcal{B})$ where the human agent (H) is 1-H-flexible with the team behavior solution $(\pi_H, \pi_R)$ and $\pi_H$ is completely accessible by robot R, i.e., $|\Pi_H|=1$ and $\mathcal{H}_\mathcal{B}=1$.
\end{definition}
Note that this is a special case of the game where we have specific assumptions on the human's preference on the team behavior and access to human's actions. While these assumptions are quite expensive to meet, they play an important role in studying Human-AI Team PbRL algorithms. Specified Orchestration provides a loose upper-bound performance by assuming away other challenging aspects of Human-AI interaction except preference elicitation, reward learning and subsequent policy learning. 
\section{Team Cooperation Human Subject Study}
\label{sec:user_study}

\subsection{Motivation} 
\label{subsec:study_motivation}
To formally investigate 1) the significance of Human Flexibility in Human-AI teaming, and, 2) the utility of Specified Orchestration in real-time scenarios, we conduct a user study (approved by our local Institutional Review Board (IRB)) using a modified Overcooked domain \cite{carroll2019utility}, as shown in Figure \ref{fig:study:grid}. In particular, we compare the scenario where the human adapts to a Helper Agent (referred to as the adaptive case) and is allowed to choose from a set of strategies to the scenario of Specified Orchestration where the human has a fixed strategy that does not change during runtime (referred to as the rigid case). We refer readers to Appendix \ref{app:user_study} for additional details on the study design.

\subsection{Evaluation Metrics}
\label{subsec:study_eval}
We use a comprehensive set of metrics to evaluate both the objective task performance and the subjective user experience in our study for both cases: Task Success, Adaptation Success and Temporal Metrics such as - "Total Time" measured the duration between the first and last action timestamps, providing insights into overall task completion and "Active Session Time" captured the duration from web-page entry to task completion or exit, excluding instructional reading time to measure task-related engagement of the users. To evaluate frustration, temporal and cognitive load on the human in the Human-AI team setting, we take subjective feedback from the users inspired by the NASA-TLX study \cite{hart2006nasa}. Here, we discuss our study findings across each of these metrics.

\subsection{Study Results}
\label{subsec:study_results}

\paragraph{Adaptation and Task Success:} Users preferred to follow their own established strategy rather than adapting to the robot's actions. Only 6 out of 20 participants adapted by picking the bowl from the passing station. Out of these, only 2 completed the task, suggesting that adaptation does not guarantee successful task completion. Interestingly, all 10 users who did not adapt to the robot's strategy and picked the bowl from their dispenser were able to successfully complete the task, implying that sticking to their own strategy, despite being different from the agent’s, led to task success. 

% We observe that users with more flexibility diverts their attention and hinders task success. The study's findings emphasize that human flexibility is an important factor to be considered for designing agents for human-AI cooperation.
\paragraph{Time spent:} In the adaptive scenario, users expended a total time of 1 minute and 33 seconds, a duration significantly longer than 46 seconds recorded in the rigid case. This discrepancy strongly suggests that users engaged in the adaptive setting invest a considerable amount of time in decision-making at each step, compelled by the need to select from multiple policies. Conversely, users in the rigid case adhered to a fixed policy, eliminating the need for continuous adaptation to the helper agent. The absence of this adaptability appears to significantly enhance efficiency, as evidenced by achieving task success within a shorter time frame. Therefore, we observe that human Flexibility has a strong impact on time taken for the execution of the task as following a predetermined policy streamlines the decision-making process. Examining active session times further underscores these findings (see Appendix \ref{subsec:app_user_study_results}). 
\paragraph{Frustration and cognitive load:} Users indicated a higher overall mental demand in the adaptive case, suggesting that adapting to the helper agent increased the cognitive workload compared to the more predetermined rigid case. Specifically, generating actions in the adaptive scenario were reported as very demanding, underscoring the challenges users faced in real-time decision-making and anticipating the actions of an adaptive agent. In assessing frustration, users reported moderate frustration levels in the adaptive case, particularly in deciding their next action, understanding the other agent's intentions, adapting to ensure task completion and relying on the other agent. Conversely, frustration levels were notably lower across all parameters in the rigid case, indicating a smoother and less frustrating experience. In summary, while the adaptive case introduced a heightened mental demand and cognitive load, participants perceived the rigid case of Specified Orchestration as more efficient and less frustrating.
\begin{figure}[!ht]
    \centering
    \includegraphics[width=0.75\linewidth]{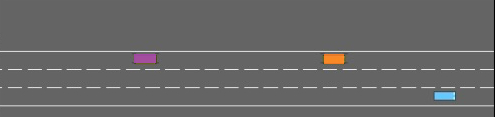}
    \caption{MA Highway domain - \textcolor{orange}{human} and the \textcolor{purple}{AI agent} cars are shown one behind the other (see Appendix \ref{app:domains}).}
    \label{fig:ma_highway_domain}
\end{figure}

\begin{figure}[!ht]
    \centering
    \includegraphics[width=1\linewidth]{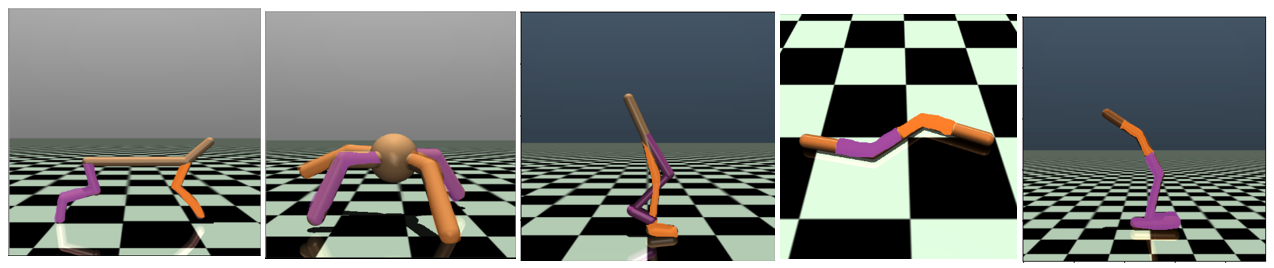}
    \caption{MA MuJoCo(\textit{L to R})- Cheetah, Ant, Walker, Swimmer, \& Hopper. The action space for the joints in each of the domains has been split between the \textcolor{orange}{human} and the \textcolor{purple}{AI agent} (see Appendix \ref{app:domains}).}
    \label{fig:ma_mujoco_domain}
\end{figure}

\section{Experiments \& Results}
\label{sec:expts_results}
\begin{figure*}[ht]
\begin{center}
\centerline{\includegraphics[width=0.98\textwidth]{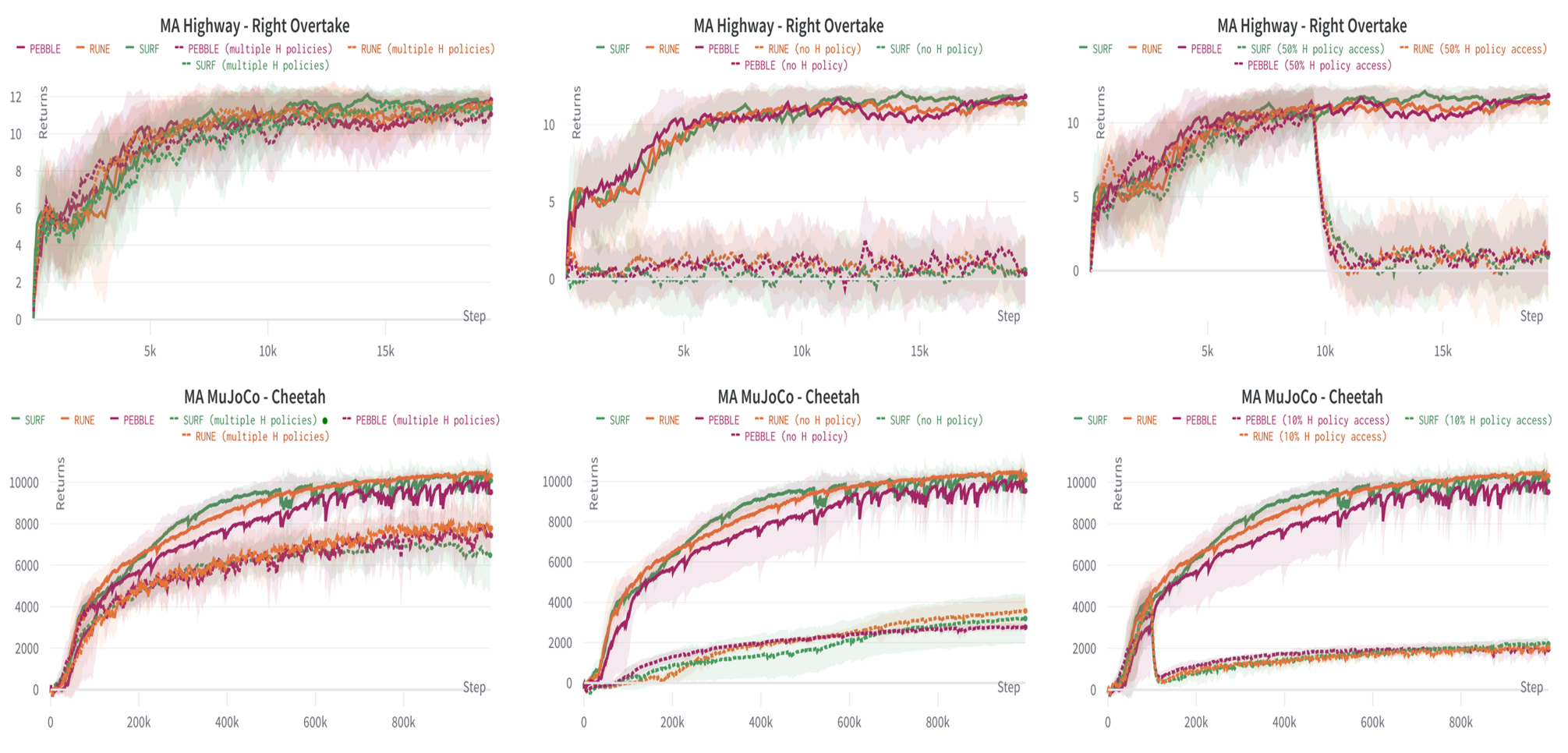}}
\caption{Learning curves on MA Highway-Right (\textit{row 1}) and MA MuJoCo - Cheetah (\textit{row 2}): comparing (\textit{from L to R}) (a) Human Flexibility on multiple $\pi_H$, with Specified Orchestration case that assumes a single $\pi_H$ and complete access to it, (b) agent's  0\% access to $\pi_H$, and (c) partial access to $\pi_H$; as measured on the episodic returns. The solid lines and shaded regions represent the mean and standard deviation, respectively, across three runs.}
\label{fig:results_image}
\end{center}
\end{figure*}

\begin{figure*}[ht]
\begin{center}
\centerline{\includegraphics[width=\textwidth]{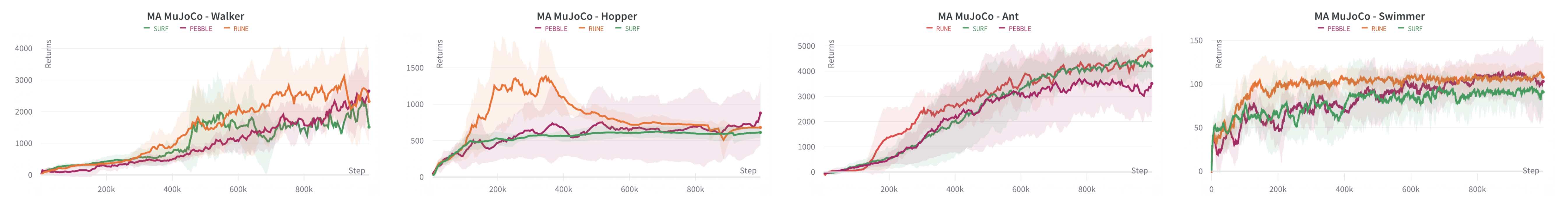}}
\caption{Learning curves on MA MuJoCo domains (from L to R) - Walker, Hopper, Ant, Swimmer: comparing PbRL algorithms under Specified Orchestration, as measured on the episodic returns. The solid lines and shaded regions represent the mean and standard deviation, respectively, across three runs.}
\label{fig:ma_mujoco_results}
\end{center}
\end{figure*}

In this section we will discuss our domain suite (Section \ref{subsec:domains}), followed by the experiment design setting (Section \ref{subsec:design_pbrl}), and our experiments evaluating multi-agent PbRL across Human Flexibility (Section \ref{subsec:flexibility_impact}), Access to Human Policy, and discussing the special case of Specified Orchestration (Section \ref{subsec:access_impact}). Further, our discussion will highlight the challenges associated with using PbRL for Human-AI team Cooperation Game and empirically answer the following questions : 
\begin{enumerate}
    \item Human Flexibility: What is the relative performance when the human partner can choose from multiple possible policies to follow, for accomplishing the team task?
    \item Access to Human Policy: How do PbRL algorithms perform with a a) budgeted and b) no access to $\pi_H$? 
    \item Specified Orchestration: What is the performance of the PbRL algorithms with maximal assumptions on access to human policy and $\epsilon=1$-H-flexible human?
\end{enumerate}
We refer the readers to the Appendix (Sections \ref{app:pebble}, \ref{app:experimentalsetup}, \ref{app:hyperparams}, and \ref{app:sec_additional_results}) for details on algorithms, the human policy and preference oracles, domain construction, hyperparameters used for the experiments and additional results, respectively.

% \begin{figure*}[ht]
% \centering 
% \subfloat[\centering MA MuJoCo - Walker]{
% \includegraphics[width=4cm, height=2.5cm]{images/results/walker_main.png}}\hfill
% \subfloat[\centering MA MuJoCo - Hopper]{
% \includegraphics[width=4cm, height=2.5cm]{images/results/hopper_main.png}}\hfill 
% \subfloat[\centering MA MuJoCo - Ant]{
% \includegraphics[width=4cm, height=2.5cm]{images/results/ant_main.png}}\hfill
% \subfloat[\centering MA MuJoCo - Swimmer]{
% \includegraphics[width=4cm, height=2.5cm]{images/results/swimmer_main.png}}
% \caption{Learning curves on MA MuJoCo domains: comparing PbRL algorithms under Specified Orchestration, as measured on the episodic returns. The solid lines and shaded regions represent the mean and standard deviation, respectively, across three runs.}
% \label{fig:ma_mujoco_results}
% \end{figure*}

\subsection{Evaluation Domains}
\label{subsec:domains}

\paragraph{MA Highway:} We design a set of three multi-agent continuous control environments on top of the Highway environment \cite{highway-env}. As shown in Figure \ref{fig:ma_highway_domain}, we construct a three-lane highway where the human car (in orange) is on the top-most lane, and the AI agent (in purple) has to overtake and merge back into the human car's lane by increasing its speed and avoiding collisions with any other car in the scene. Analogous to the Overcooked domain, MA Highway also does not require forced cooperation for task completion.

\paragraph{MA MuJoCo:} Following prior works in constructing multi-agent environments \cite{peng2021facmac} in continuous control domains and PbRL setups \cite{bpref}, we modify the MuJoCo-based locomotion environments \cite{todorov2012mujoco} for the human-AI teaming tasks which necessitate cooperation. We construct multi-agent settings for five environments, as shown in Figure \ref{fig:ma_mujoco_domain}.

\subsection{Adapting PbRL for Human-AI Team setting}
\label{subsec:design_pbrl}
For our experiments, we assume a human policy set $|\Pi|$ from which the human agent can select a single (or multiple) policies for themselves, and the AI agent has the access to the human policy during training with a budget $\mathcal{H}_\mathcal{B}$. Following prior works \cite{fcp,hsp}, when a new episode begins, we uniformly sample a human policy $\pi_H$ from $|\Pi|$. Note, if the human only has a single policy, i.e., in the case of Specified Orchestration, then we use the same $\pi_H$ for every episode. During the policy learning phase for the AI agent if the agent has remaining budget to request actions from human policy, we sample $a_H$ from $\pi_H$, or else, the AI agent assumes a random action on behalf of the human. Again, in the case of Specified Orchestration, the budget $\mathcal{H}_\mathcal{B}$ allows the AI agent to access human actions till the end of the training (see Algorithm \ref{alg:pebble_algorithm}). 

% As per our re-interpretation of the Human-AI team PbRL Cooperation game in Section \ref{subsec:prelim_markov_game}, this is equivalent to the assumption that human is acting in the shared world with the agent. 

\subsection{Impact of Human Flexibility}
\label{subsec:flexibility_impact}

To study the impact of \textit{Human Flexibility} (our Q1) on PbRL performance we operate under the assumption that agent has complete access to the human policy (i.e. human executes their policy during agent's training). We assume that for each episode of learning, the human agent can sample 1-out-of-N (3, in our case) policies to provide the actions to the AI agent. From the results shown in the first column of Figure \ref{fig:results_image}, we note that the performance comparatively drops as in the case where we assumed a single $\pi_H$ and complete access to it. The decline in performance is amplified in the more complex case of MA MuJoCo Cheetah where cooperation is forced amongst the two agents. We conclude that the team performance is the highest in the case of Specified Orchestration, where the human agent has a single $\pi_H$ and the AI agent has complete access to the same.
\paragraph{\textbf{Observed Challenges:}} From the experiments with multiple $\pi_H$ available to the AI agent during the training phase, we conclude what while it is a cognitively challenging task for a human agent to compute multiple policies and also provide feedback in this case, reward learning is also negatively impacted as seen from our empirical analysis.

\subsection{Impact of Access to Human Policy}
\label{subsec:access_impact}
\paragraph{\textbf{No Access to Human policy:}}
In reference to Question (2), we report the performance of PEBBLE \cite{pebble}, RUNE \cite{rune}, and SURF \cite{surf} on the three MA Highway and MA MuJoCo - Cheetah domains in second column of Figure \ref{fig:results_image}. None of the algorithms have any access to $\pi_H$, and thus, have to `imagine', i.e., assume random actions for the human agent. In all the four domains and three algorithms, we note that the performance is significantly worse as compared to the case when the AI agent has complete access to $\pi_H$. Utilizing explicit modeling of human policy is a natural next step and an important future work, but it is beyond the scope of this work.
\paragraph{\textbf{Restricting Access to Human Policy:}}
Furthermore, we first investigate the case where we restrict the AI agent's budget for accessing the human policy $\pi_H$ to 50\% for MA Highway domain, and 10\% for MA MuJoCo - Cheetah (third column of Figure \ref{fig:results_image}). We allocate this entire budget in the first half of the training. For example, in the case of MA MuJoCo - Cheetah domain, the AI agent has access to $\pi_H$ up to 100k steps, and assumes random actions sampled for the remaining training steps. Expectedly, we note that the performance plummets significantly as soon as the human agent's actions are not available for all the domains.
\paragraph{\textbf{Specified Orchestration - complete access to $\pi_H$:}}
Finally, to answer Question (3) supporting our discussion on Specified Orchestration, we report the performance of all three PbRL algorithms in Figures \ref{fig:results_image} and \ref{fig:ma_mujoco_results}. We note from each of the plots, that all three algorithms show the optimal returns. From Figure \ref{fig:ma_mujoco_results}, we also note that RUNE consistently performs better than SURF and PEBBLE in these multi-agent settings. 
\paragraph{\textbf{Observed Challenges:}} From the experiments with 0\% and partially restricted access to the human partner policy, we observe that reward learning is severely impacted. Since Specified Orchestration assumes maximal information on the human partner's policy and preferences, the team performance is optimal. This highlights the potential of using Specified Orchestration as a means to obtain a loose upper bound on the Human-AI team PbRL Cooperation Game performance. Further, unless the AI agent explicitly models human policy (and preferences) with restricted access, both reward and policy learning are challenging problems.

\section{Conclusion}
\label{sec:conclusion}
We perform a first exploration of Preference-based Reinforcement Learning for Human-AI team setup. We introduce a Human-AI PbRL Cooperation game and argue across two critical dimensions, namely - \textit{Human Flexibility} and AI agent's \textit{Access to Human Policy}. In our setup, both the agents are rewarded according to human's reward function for the team comprising of human's objectives for the team, sub-objectives for themselves and the robot, and human preferences on team behavior. To this end, we evaluate PbRL algorithms on our domain suite consisting of simpler Highway and more complex robot locomotion domains. From our experiments across the different cases of \textit{Human Flexibility} and AI agent's \textit{Access to Human Policy}, we conclude that Specified Orchestration, while assuming the maximal amount of information on the human, shows notable performance across all domains. However, this can also be seen as one of the key limitations as it may not always be the case that the human can easily compute and provide their policy to the AI agent (or be available to act during the training procedure of the AI agent). Secondly, while this approach allows for optimal team performance of an AI agent with a specific human, it may be hard to generalize the learnt reward function (and hence, the learnt policy) to multiple human partners. Thirdly, even the case of Specified Orchestration assumes the human agent to act rationally and within the confines of pre-defined strategies, which may not always hold true in real-world scenarios where human behavior can be predictable. Lastly, as in the case of single-agent PbRL works where human is just an observer, the problem of expensive human interaction with the AI agent is carried over to our setting. To conclude, our work provides researchers with a first exploration of PbRL beyond single-agent setup, towards human-AI teaming.

\bibliography{aaai25}

\clearpage
\newpage
\appendix

\section{PbRL Algorithm}
\label{app:pebble}
We present our modified PbRL algorithm for the human-AI team setting, which uses the PEBBLE algorithm \cite{pebble} as a backbone. Note, we make the same modifications for experiments using RUNE \cite{rune} and SURF \cite{surf} as these algorithms also use PEBBLE as their PbRL backbone. We do not require any change to the model architectures (including reward approximator and policy approximator). The updates are highlighted in red (see Algorithm \ref{alg:pebble_algorithm}) and can be easily incorporated into any online single agent PbRL algorithm.

Following past works in Preference based Reinforcement Learning,  we use a simulated oracular human policy to provide the actions on behalf of the human agent. As suggested in \cite{bpref} we use a scripted preference labeling function that provides preference labels on behalf of the human agent to the PbRL agent. We train SAC \cite{haarnoja2018soft} by assuming centralized controller for human and AI agent using underlying task reward available with the environment package for MuJoCo \cite{openaigym} and handcrafted reward function for Highway. We take the actions corresponding to the joint angles under control of the human as human actions and rest for the AI agent's actions expected by the human as part of the compatible policy pair. Preference labels are generated based on hand-crafted rewards for MA Highway and environment packaged rewards (which serve as team reward) for MA MuJoCo. We refer the readers to the appendix for further details. 

\begin{algorithm}[ht]
   \caption{Adapting Single Agent PbRL for Human-AI Team}
   \label{alg:pebble_algorithm}
\begin{algorithmic}
   \STATE {\bfseries Input:} feedback frequency $K$, \# queries per feedback session $M$, \textcolor{red}{human policy set: $|\Pi|$, agent access budget: $b$}
   \STATE Initialize parameters of $Q_\theta$ and $\hat{r}_\psi$
   \STATE Initialize a dataset of preferences $D_h \leftarrow \varnothing$
   \STATE // EXPLORATION PHASE
   \STATE $D_\tau, \pi_\phi \leftarrow $ \texttt{EXPLORE()} in \cite{pebble}
   \STATE POLICY LEARNING PHASE
   \FOR{each iteration}
    \STATE // REWARD LEARNING PHASE
    \IF{\textcolor{red}{new episode}}
        \STATE \textcolor{red}{choose human policy $\pi_H \in |\Pi|$}
    \ENDIF
    \IF{iteration \% $K$ == 0}
        \FOR{m in 1...$M$}
            \STATE $(\sigma^0, \sigma^1) \sim$ \texttt{SAMPLE()} in \cite{pebble}
            \STATE Query \textcolor{red}{human partner} for $y$
            % \STATE Store preference $D_h \leftarrow D_h \union \{(\sigma^0,\sigma^1,y)\}$
        \ENDFOR
        \FOR{each gradient step}
            \STATE Sample mini-batch $\{{(\sigma^0,\sigma^1,y)}_j\}^{D_h}_{j=1} \sim D_h$
        \ENDFOR
        \STATE Relabel entire replay buffer $D_{\tau}$ using $\hat{r}_\psi$
    \ENDIF
    \STATE REINFORCEMENT LEARNING PHASE
   \FOR{each time-step $t$}
    \IF{\textcolor{red}{time-step $t \leq$ budget $b$}}
        \STATE \textcolor{red}{sample $a_h \sim \pi_H$}
    \ELSE
        \STATE \textcolor{red}{randomly sample $a_h \sim \mathcal{A}^H$ }    
    \ENDIF
    \STATE Collect $s_{t+1}$ by taking $a_t \sim \pi_\phi(a_t|s_t)$ and \textcolor{red}{$a_H$}
    \STATE Store transitions $D_\tau \leftarrow D_\tau \{(s_t,a_t,\textcolor{red}{a_h},s_{t+1},\hat{r}_\psi(s_t))\}$
   \ENDFOR
   \FOR{each gradient step}
        \STATE Sample random mini-batch $\{{\tau_j}\}^{D_\tau}_{j=1} \sim D_\tau$
        \STATE Optimize $L^{SAC}_{critic}$ and $L^{SAC}_{actor}$ w.r.t. $\theta$ and $\phi$, respectively \cite{pebble}
   \ENDFOR
   \ENDFOR
\end{algorithmic}
\end{algorithm}

\subsection{PEBBLE Algorithm}

PEBBLE is a PbRL algorithm that comprises of two key elements: pre-training and relabeling experience buffer. To gather a wide range of experiences, PEBBLE starts by using intrinsic motivation \cite{chentanez2004intrinsically,barto2013intrinsic,abel2021expressivity,schmidhuber2010formal} to pre-train the policy, which optimizes the policy to increase the state entropy in order to explore the environment effectively. Afterwards, PEBBLE uses the SAC algorithm, a state-of-the-art off-policy RL algorithm, to further train the policy. To ensure stability in the learning process, PEBBLE relabels all experiences in the buffer when the reward model is updated.

\subsection{Modified PEBBLE Algorithm for varying \textit{Access to Human Policy} experiments}

Please refer to Algorithm \ref{alg:pebble_algorithm_access} for more details.

\begin{algorithm}[ht]
   \caption{Adapting Single Agent PbRL for Human-AI Team}
   \label{alg:pebble_algorithm_access}
\begin{algorithmic}
   \STATE {\bfseries Input:} feedback frequency $K$, \# queries per feedback session $M$, \textcolor{red}{human policy set: $|\Pi| = {\pi_H}$, agent access budget: $b \sim [0, 1]$}
   \STATE Initialize parameters of $Q_\theta$ and $\hat{r}_\psi$
   \STATE Initialize a dataset of preferences $D_h \leftarrow \varnothing$
   \STATE // EXPLORATION PHASE
   \STATE $D_\tau, \pi_\phi \leftarrow $ \texttt{EXPLORE()} in \cite{pebble}
   \STATE POLICY LEARNING PHASE
   \FOR{each iteration}
    \STATE // REWARD LEARNING PHASE
    \IF{\textcolor{red}{new episode}}
        \STATE \textcolor{red}{choose human policy $\pi_H$}
    \ENDIF
    \IF{iteration \% $K$ == 0}
        \FOR{m in 1...$M$}
            \STATE $(\sigma^0, \sigma^1) \sim$ \texttt{SAMPLE()} in \cite{pebble}
            \STATE Query \textcolor{red}{human partner} for $y$
            % \STATE Store preference $D_h \leftarrow D_h \union \{(\sigma^0,\sigma^1,y)\}$
        \ENDFOR
        \FOR{each gradient step}
            \STATE Sample mini-batch $\{{(\sigma^0,\sigma^1,y)}_j\}^{D_h}_{j=1} \sim D_h$
        \ENDFOR
        \STATE Relabel entire replay buffer $D_{\tau}$ using $\hat{r}_\psi$
    \ENDIF
    \STATE REINFORCEMENT LEARNING PHASE
   \FOR{each time-step $t$}
    \IF{\textcolor{red}{time-step $t \leq$ budget $b$}}
        \STATE \textcolor{red}{sample $a_h \sim \pi_H$}
    \ELSE
        \STATE \textcolor{red}{randomly sample $a_h \sim \mathcal{A}^H$ }    
    \ENDIF
    \STATE Collect $s_{t+1}$ by taking $a_t \sim \pi_\phi(a_t|s_t)$ and \textcolor{red}{$a_H$}
    \STATE Store transitions $D_\tau \leftarrow D_\tau \{(s_t,a_t,\textcolor{red}{a_h},s_{t+1},\hat{r}_\psi(s_t))\}$
   \ENDFOR
   \FOR{each gradient step}
        \STATE Sample random mini-batch $\{{\tau_j}\}^{D_\tau}_{j=1} \sim D_\tau$
        \STATE Optimize $L^{SAC}_{critic}$ and $L^{SAC}_{actor}$ w.r.t. $\theta$ and $\phi$, respectively \cite{pebble}
   \ENDFOR
   \ENDFOR
\end{algorithmic}
\end{algorithm}

\subsection{Modified PEBBLE Algorithm for \textit{Human Flexibility} experiments}

Please refer to Algorithm \ref{alg:pebble_algorithm_flexibility} for more details.

\begin{algorithm}[ht]
   \caption{Adapting Single Agent PbRL for Human-AI Team}
   \label{alg:pebble_algorithm_flexibility}
\begin{algorithmic}
   \STATE {\bfseries Input:} feedback frequency $K$, \# queries per feedback session $M$, \textcolor{red}{human policy set: $|\Pi|$, agent access budget: $b = {1}$}
   \STATE Initialize parameters of $Q_\theta$ and $\hat{r}_\psi$
   \STATE Initialize a dataset of preferences $D_h \leftarrow \varnothing$
   \STATE // EXPLORATION PHASE
   \STATE $D_\tau, \pi_\phi \leftarrow $ \texttt{EXPLORE()} in \cite{pebble}
   \STATE POLICY LEARNING PHASE
   \FOR{each iteration}
    \STATE // REWARD LEARNING PHASE
    \IF{\textcolor{red}{new episode}}
        \STATE \textcolor{red}{choose human policy $\pi_H \in |\Pi|$}
    \ENDIF
    \IF{iteration \% $K$ == 0}
        \FOR{m in 1...$M$}
            \STATE $(\sigma^0, \sigma^1) \sim$ \texttt{SAMPLE()} in \cite{pebble}
            \STATE Query \textcolor{red}{human partner} for $y$
            % \STATE Store preference $D_h \leftarrow D_h \union \{(\sigma^0,\sigma^1,y)\}$
        \ENDFOR
        \FOR{each gradient step}
            \STATE Sample mini-batch $\{{(\sigma^0,\sigma^1,y)}_j\}^{D_h}_{j=1} \sim D_h$
        \ENDFOR
        \STATE Relabel entire replay buffer $D_{\tau}$ using $\hat{r}_\psi$
    \ENDIF
    \STATE REINFORCEMENT LEARNING PHASE
   \FOR{each time-step $t$}
    % \IF{\textcolor{red}{time-step $t \leq$ budget $b$}}
    %     \STATE \textcolor{red}{sample $a_h \sim \pi_H$}
    % \ELSE
    %     \STATE \textcolor{red}{randomly sample $a_h \sim \mathcal{A}^H$ }    
    % \ENDIF
    \STATE \textcolor{red}{sample $a_h \sim \pi_H$}
    \STATE Collect $s_{t+1}$ by taking $a_t \sim \pi_\phi(a_t|s_t)$ and \textcolor{red}{$a_H$}
    \STATE Store transitions $D_\tau \leftarrow D_\tau \{(s_t,a_t,\textcolor{red}{a_h},s_{t+1},\hat{r}_\psi(s_t))\}$
   \ENDFOR
   \FOR{each gradient step}
        \STATE Sample random mini-batch $\{{\tau_j}\}^{D_\tau}_{j=1} \sim D_\tau$
        \STATE Optimize $L^{SAC}_{critic}$ and $L^{SAC}_{actor}$ w.r.t. $\theta$ and $\phi$, respectively \cite{pebble}
   \ENDFOR
   \ENDFOR
\end{algorithmic}
\end{algorithm}

\clearpage
\newpage

\section{User Study Details}
\label{app:user_study}

% \textcolor{red}{
% \begin{enumerate}
%     \item add IRB approval 
%     \item demographics of participants - Table and 2 lines of description
%     \item Study design - figures from the study - screenshots of all pages, instruction videos and trial game, instructions for Case 1 and 2, list of all questions given at the end of the study and choice of Likert Scale
%     \item expand results - provide statistics for the two cases in a table (all time values), add corresponding text, also add results and plots for question/answers and discuss in text drawing conclusions.
% \end{enumerate}
% }
% \section{IRB Approval}
All participants were provided with a detailed information page explaining the study's instructions, expected duration and payment details. Participants were assured that their involvement in the study adheres to strict confidentiality measures, aligning with IRB guidelines (code redacted for anonymity). By continuing with the study, participants consented to the secure recording of their data for research purposes.
\begin{figure}[hbt!]
    \centering
    \includegraphics[width=1\linewidth]{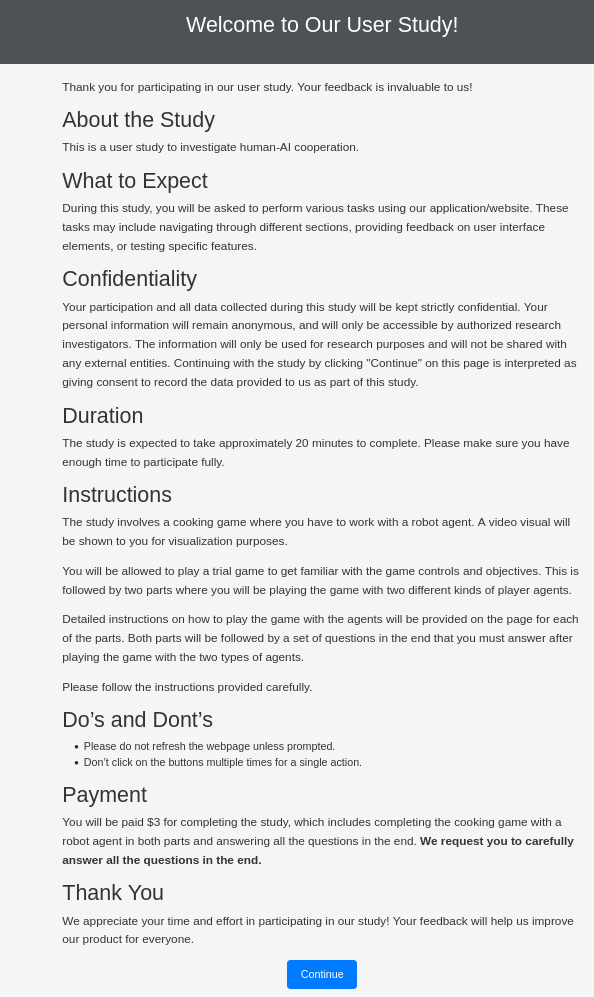}
    \caption{Welcome Page for the Use Study}
    \label{fig:welcome}
\end{figure}

\begin{figure}[!hbt]
    \centering
    \includegraphics[width=1\linewidth]{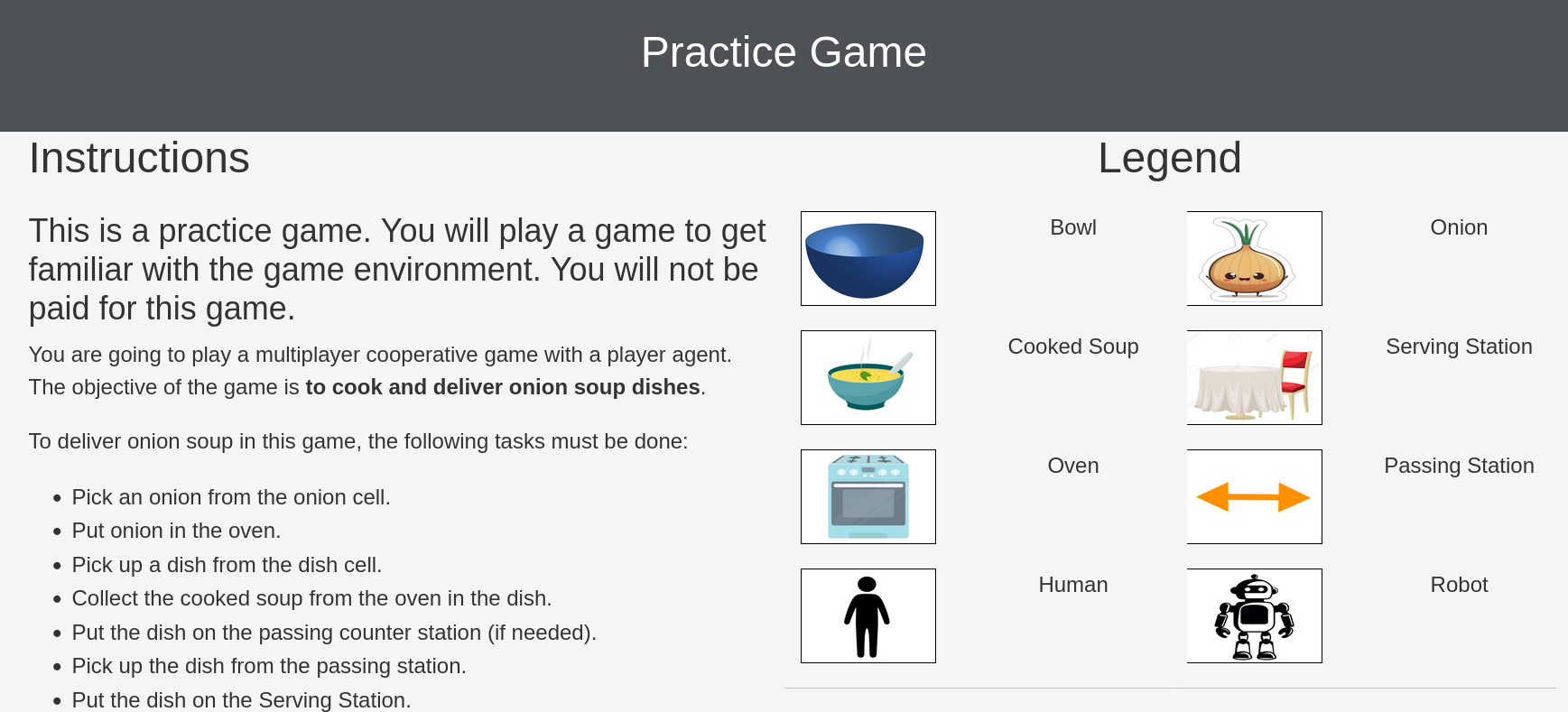}
    \caption{Description of the Task Objectives, Controls and Legend}
    \label{fig:instructions}
\end{figure}
\begin{figure}[!hbt]
    \centering
    \includegraphics[width=1\linewidth]{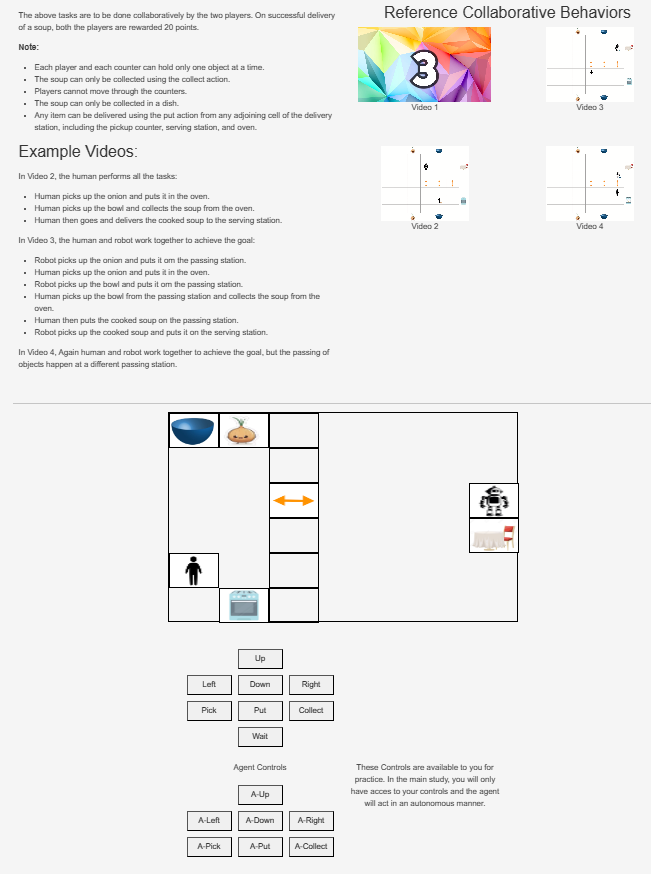}
    \caption{Example Videos to Illustrate Multiple Joint Policies for Task Success and practice game with controls.}
    \label{fig:videos-game}
\end{figure}
\begin{figure}[!hbt]
    \centering
    \includegraphics[width=1\linewidth]{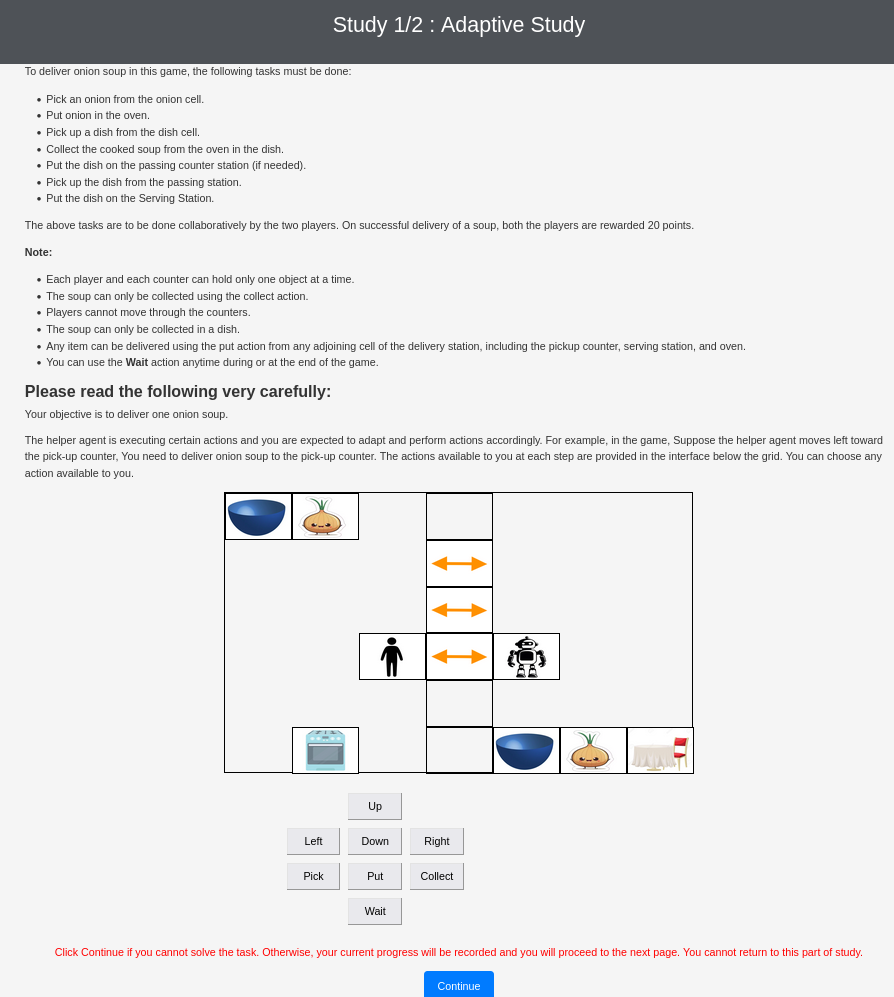}
    \caption{Instructions Provided for Adaptive Case}
    \label{fig:adapt}
\end{figure}
\begin{figure}[!ht]
    \centering
    \includegraphics[width=1\linewidth]{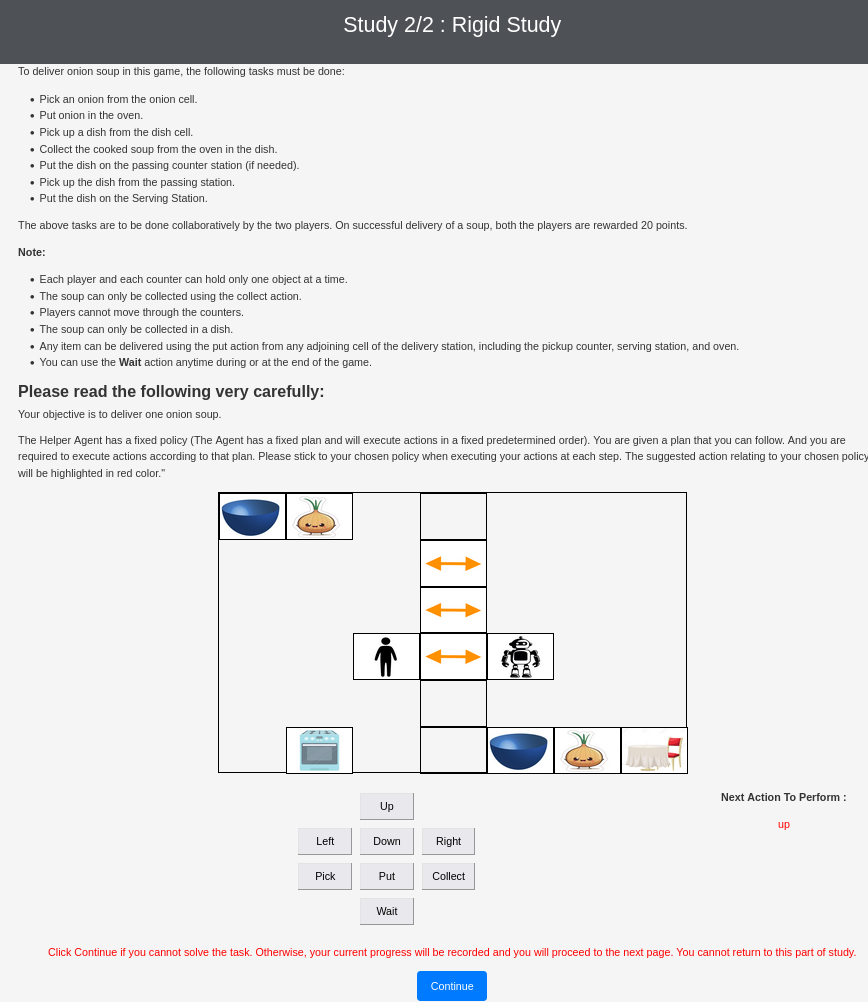}
    \caption{Instructions Provided for Specified Orchestration Case}
    \label{fig:rigid}
\end{figure}
\subsection{Study Design}
Inspired by the Overcooked domain \cite{carroll2019utility}, we implement a multi-player cooperative game where participants engage with a Helper Agent in an environment featuring Onion dispensers, Dish dispensers, Oven, a Serving station, and empty counters, with the goal of cooking and delivering onion soup dishes. The dynamic game-play allows players to move within the environment, executing actions such as picking, placing, and collecting objects. The process of preparing onion soup involves picking onions, placing them in the oven, and collecting the soup in a dish. Successful delivery of the soup to the pickup counter rewards both players with 20 points. The study introduces different treatments within the same control group, varying the sizes of policy sets to observe the impact on subjective experience and team performance. Participants initially come across a page containing game instructions. This page provides details on game controls, task objectives, and a legend explaining the symbols and their corresponding meanings in the game environment as shown in Figure \ref{fig:instructions}. Next, participants are presented with demo videos illustrating examples of how the task can be accomplished by the two agents with different levels of collaboration and allowed to test out the controls in the game environment (Figure \ref{fig:videos-game}). In the first treatment, participants have the flexibility to select from a large set of policies, adapting to the Helper Agent's dynamic execution of actions during the game (Figure \ref{fig:adapt}). The instructions provided to the agents in the webpage are - "The helper agent is executing certain actions and you are expected to adapt and perform actions accordingly. For example, in the game, Suppose the helper agent moves left toward the pick-up counter, You need to deliver onion soup to the pick-up counter. The actions available to you at each step are provided in the interface below the grid. You can choose any action available to you". The second treatment introduces a fixed policy for the Helper Agent, wherein players receive a predetermined plan to follow (Figure \ref{fig:rigid}). This is the case of Specified Orchestration where the set of feasible strategies is singleton for the players. Players are required to adhere to their chosen policy at each step, helping investigate how different policy set sizes influence the overall game-play experience and team dynamics. The following instructions are provided to the participants - "The Helper Agent has a fixed policy (The Agent has a fixed plan and will execute actions in a fixed predetermined order). You are given a plan that you can follow. And you are required to execute actions according to that plan. Please stick to your chosen policy when executing your actions at each step. The suggested action relating to your chosen policy will be highlighted in red color". 

\label{subsec:study_design}

\subsection{Additional Results}
\label{subsec:app_user_study_results}
\paragraph{Demographics Statistics:}
The user study involves a diverse demographic, with participants averaging 35 years of age and exhibiting a standard deviation of 12.78. The gender distribution showcases 6 male, 13 female, and 1 participant identifying as 'Other,' ensuring a varied and representative sample for the study.
\begin{table}[!hbt]
\centering
\caption{User Study Demographics Statistics: The User study respondents consisted of 20 Participants with the following age and gender distributions.}
\label{tab:user_study_demographics}
\begin{tabular}{ccc}
\hline
\textbf{Age Statistics} \\ \hline
Average                                           & Standard Deviation\\
$35$ & $12.78$ \\
\hline
\textbf{Gender Statistics} \\ \hline
Male & Female & Other \\
$6$ & $13$ & $1$ \\
\end{tabular}
\end{table}
\paragraph{Session Time:} In the adaptive case, the total session time was 2 minutes and 1 second, surpassing 1 minute and 16 seconds recorded in the rigid (Specified Orchestration) case. Interestingly, even though users in the rigid case invested time upfront in policy selection before the run begun, the total active session time (the duration when users actively engaged in the task) was markedly shorter than that observed in the adaptive case. This reinforces the argument for Specified Orchestration, where the human communicates it's selected policy to the agent, and maintains a single policy in their set of feasible team strategies. 

We also conducted the Student's t-test to compare the total session time for all participants in the rigid vs. adaptive case. Our null hypothesis is: there is no difference between the time taken by participants to finish the task in the rigid (Specified Orchestration) case and the adaptive case. The t-test revealed that there is a statistically significant difference in the total session time (p = 0.043073 $<$ 0.05) as well as the total active session time (p = 0.02953 $<$ 0.05) for participants in the rigid case and adaptive case. Thus, we reject the null hypothesis.
\begin{table}[ht]
\centering
\caption{User Study Time Metrics}
\label{tab:user_study_time_metrics}
\resizebox{\columnwidth}{!}{%
\begin{tabular}{ccc}
\hline
\textbf{Time Metrics} & \textbf{Case} & \textbf{Time (s)} \\
\hline
\multirow{2}{12em}{Average Total Session Time} & Specified Orchestration & 81 \\
& Adaptive Case & 112 \\ \hline
\multirow{2}{12em}{Average Active Session Time} & Specified Orchestration & 52 \\
& Adaptive Case & 82 \\ \hline

\end{tabular}%
}
\end{table}

\paragraph{Subjective Metrics:}
At the end of the user study, we administer a questionnaire to assess subjective aspects of the user experience for the game, specifically focusing on mental demand, time pressure, task efficiency, and frustration for both cases. These factors directly contribute to the evaluation of cognitive load, temporal load, task performance and frustration as shown in Table \ref{tab:user_study_stats}.

\begin{figure}[ht]
\centering
\includegraphics[width=1\linewidth]{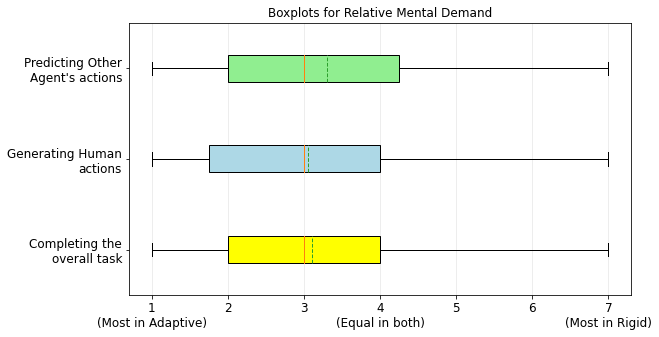}
\caption{Boxplots representing Cognitive Load factors that shows responses from the User Study evaluating relative mental demand on a Likert Scale of 1 to 7 for (i) Completing the Overall task. (ii) Generating their actions. (iii) Predicting Other Agent's actions. The green and orange lines within the boxes represent the mean and median respectively.}
\label{fig:study_cognitive_load}
%\end{center}
\end{figure}

\begin{figure}[ht]
\centering
\includegraphics[width=1\linewidth]{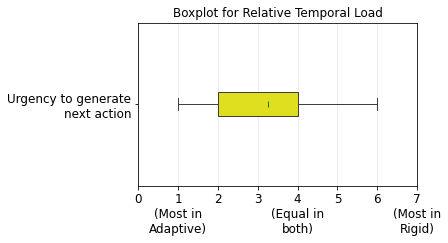}
\caption{Boxplots representing Temporal Load responses from the User Study evaluating relative task pressure on a Likert Scale of 1 to 7. The green line within the box represents the mean.}
\label{fig:study_temporal_load}
%\end{center}
\end{figure}

\begin{figure}[ht]
\centering
\includegraphics[width=1\linewidth]{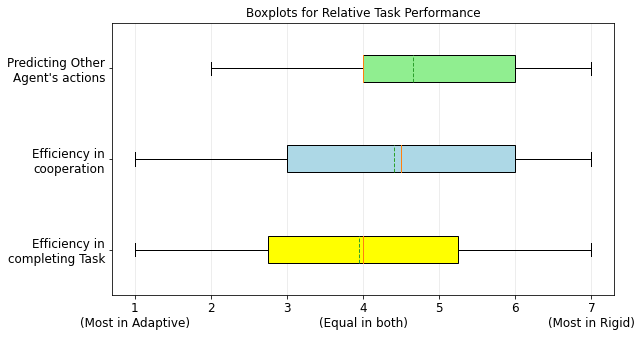}
\caption{Boxplots representing Task Performance factors that show subjective responses from the User Study evaluating relative task efficiency on a Likert Scale of 1 to 7 for (i) Efficiency in Task Completion. (ii) Efficiency in Cooperation. (iii) Predicting Other Agent's actions. The green and orange lines within the boxes represent the mean and median, respectively.}
\label{fig:study_task_performance}
%\end{center}
\end{figure}

\begin{figure}[ht]
\centering
\includegraphics[width=1\linewidth]{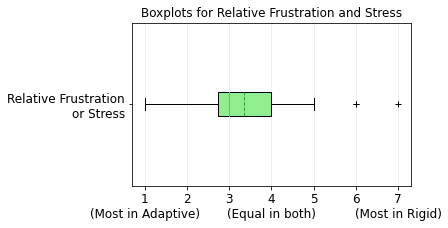}
\caption{Boxplots representing Relative Frustration and Stress that show subjective responses from the User Study on a Likert Scale of 1 to 7. The green and orange lines within the box represent the mean and median, respectively.}
\label{fig:study_relative_frustration}
%\end{center}
\end{figure}

\begin{figure}[ht]
\centering
\includegraphics[width=1\linewidth]{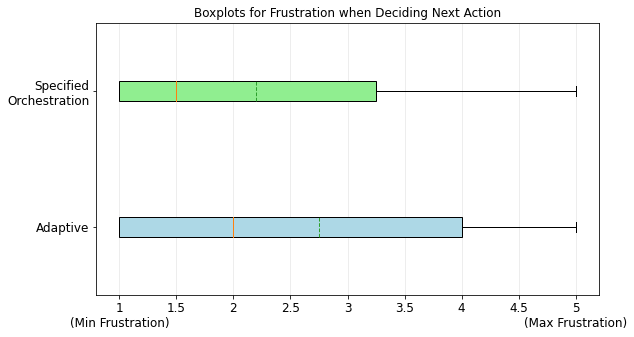}
\caption{Boxplots comparing User Frustration levels when deciding their next action in Specified Orchestration and Adaptive Cases using a Likert Scale of 1 to 5 as shown. The green and orange lines within the boxes represent the mean and median, respectively.}
\label{fig:study_Frustration_1}
%\end{center}
\end{figure}

\begin{figure}[ht]
\centering
\includegraphics[width=1\linewidth]{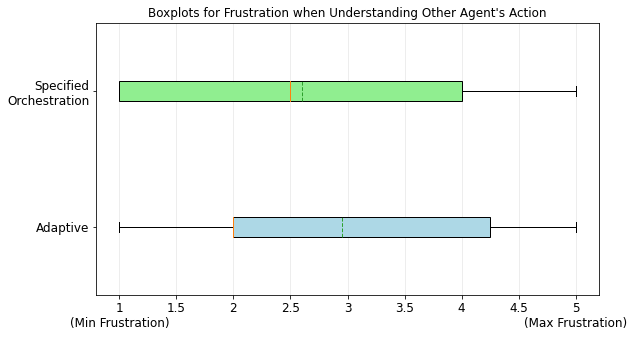}
\caption{Boxplots comparing User Frustration levels in understanding other agent's intention for Specified Orchestration and Adaptive Cases using a Likert Scale of 1 to 5 as shown. The green and orange lines within the boxes represent the mean and median, respectively.}
\label{fig:study_Frustration_2}
%\end{center}
\end{figure}

\begin{figure}[ht]
\centering
\includegraphics[width=1\linewidth]{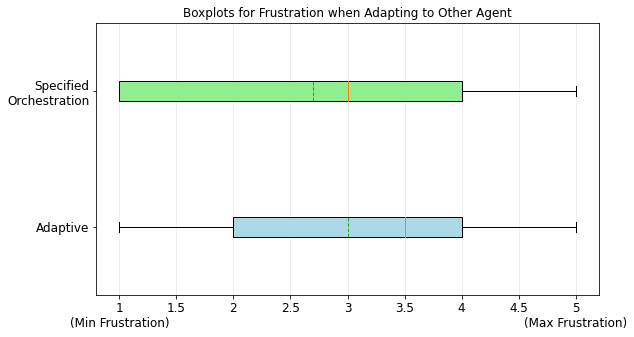}
\caption{Boxplots comparing User Frustration levels when adapting to the other agent in Specified Orchestration and Adaptive Cases using a Likert Scale of 1 to 5 as shown. The green and orange lines within the boxes represent the mean and median, respectively.}
\label{fig:study_Frustration_3}
%\end{center}
\end{figure}

\begin{figure}[ht]
\centering
\includegraphics[width=1\linewidth]{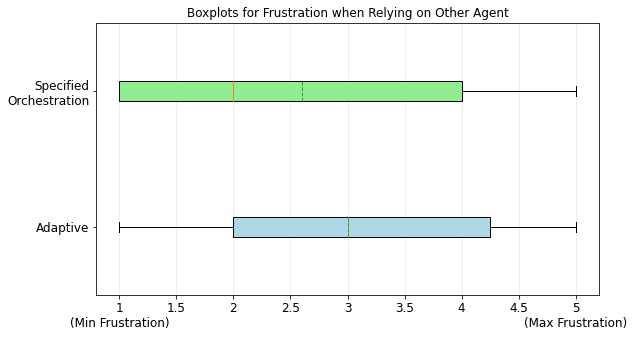}
\caption{Boxplots comparing User Frustration levels when relying on the other agent to ensure task completion in Specified Orchestration and Adaptive Cases using a Likert Scale of 1 to 5 as shown. The green and orange lines within the boxes represent the mean and median respectively.}
\label{fig:study_Frustration_4}
%\end{center}
\end{figure}

Questionnaire:
\begin{enumerate}
    \item Relative Mental Demand:
        \begin{enumerate}
        \item How mentally demanding was completing the overall task, relatively in the two cases?
        \item How mentally demanding was generating your actions, relatively in the two cases?
        \item How mentally demanding was predicting the actions of the other agent at every step, relatively in the two cases?
        \end{enumerate}
    \item Relative Time Pressure: How hurried or rushed did you feel to come up with your next action at every step, relatively in the two cases?
    \item Relative Efficiency:
        \begin{enumerate}
            \item How efficient do you think you were in accomplishing the task, relatively in the two cases?
            \item How efficient do you think you were in cooperating with the other agent to complete the task, relatively in the two cases? (Cooperating here refers to the ease of working together without any direct communication with your partner.)
            \item How efficient do you think you were in predicting what the other agent is going to do at every step, relatively in the two cases?
        \end{enumerate}
    \item Relative Frustration: How insecure, discouraged, irritated, stressed, and annoyed were you, relatively in the two cases?
    \item Frustration for Adaptive Case: On a scale of 1-5, rate your frustration for the following four parameters in Case 1:
        \begin{enumerate}
            \item Deciding your next action:
            \item Understanding the other agent’s next action or intention:
            \item Adapting to the other agent to ensure task completion:
            \item Relying on the other agent to ensure task completion:
        \end{enumerate}
    \item Frustration for Rigid Case: On a scale of 1-5, rate your frustration for the following four parameters in Case 2:
    \begin{enumerate}
        \item Deciding your next action:
        \item Understanding the other agent's next action or intention:
        \item Adapting to the other agent to ensure task completion:
        \item Relying on the other agent to ensure task completion:
    \end{enumerate}
\end{enumerate}
To gauge relative user preferences between the adaptive and rigid (Specified Orchestration) scenarios, participants provide input on a Likert scale ranging from 1 ('Most in Case 1') to 7 ('Most in Case 2'), with 4 indicating equality in both cases. Additionally, frustration levels are evaluated on a separate Likert scale, ranging from 1 ('Minimum frustration') to 5 ('Maximum frustration'), aiming to comprehensively understand and compare the frustration experience by the users in both cases. Statistics for participants' responses to the questions are presented in the Figure \ref{fig:subj-result} and visualized in Figure \ref{fig:study_cognitive_load}, Figure \ref{fig:study_temporal_load}, Figure \ref{fig:study_task_performance}, Figure \ref{fig:study_relative_frustration}, Figure \ref{fig:study_Frustration_1}, Figure \ref{fig:study_Frustration_2}, Figure \ref{fig:study_Frustration_3} and Figure \ref{fig:study_Frustration_4} to offer a comprehensive overview of the subjective aspects, including mental demand, time pressure, task efficiency, and frustration.
The provided statistical plots offer a comprehensive view of the participants' experiences across various dimensions including the metrics from Table \ref{tab:user_study_stats}. Participants reported a significantly higher mental demand in the adaptive case as compared to the specified orchestration case for completing overall task, generating their actions and predicting other agents' actions. The urgency to generate the next action indicated a notable time pressure for the adaptive case, suggesting a pressure to generate actions in real-time for users in this case. Regarding task efficiency, respondents perceived higher efficiency in completing tasks and cooperation in the Specified Orchestration (rigid) case, where predictive abilities regarding the other agent's actions were notably high. Frustration levels were higher in the adaptive case, with the adaptive case showing higher frustration in decision-making, understanding the other agent's intention, and reliance on the other agent. Conversely, specified orchestration demonstrated lower frustration levels across these parameters.

\begin{table*}[ht]
%\begin{adjustbox}{width=\columnwidth,center}
\centering
\caption{User Study Descriptive Statistics for Evaluation}
\label{tab:user_study_stats}
\resizebox{\textwidth}{!}{%
\begin{tabular}{c|c|c|ccc}
\hline
\textbf{Evaluation Factors} & \textbf{Study sub-factors} & \textbf{Question Focus} & \textbf{Mean} & \textbf{Standard Deviation} & \textbf{Median} \\ \hline
\multirow{3}{7em}{Cognitive Load} & \multirow{3}{11em}{Relative Mental Demand} & Completing the overall task & 3.1 &	1.74 & 3 \\ 
& & Generating your actions& 3.05	& 1.85 & 3 \\ 
& & Predicting Other Agent's actions	& 3.3 & 1.72 & 3
 \\ \hline
\multirow{1}{7em}{Temporal Load} & \multirow{1}{11em}{Relative Time Pressure} & Urgency to generate next action & 3.25 & 1.55 & 4 \\ \hline
\multirow{3}{7em}{Task Performance} & \multirow{3}{11em}{Relative Task Efficiency} & Efficiency in completing Task & 3.95 & 2.09 & 4 \\ 
& & Efficiency in cooperation & 4.4 & 1.73 & 4.5 \\ 
& & Predicting Other Agent's actions & 4.65 & 1.6 & 4 \\ \hline
\multirow{9}{7em}{Frustration} & \multirow{1}{11em}{Relative Frustration/ Stress} & Relative Frustration or Stress & 3.35 & 1.73 & 3 \\ \cline{2-6}
& \multirow{4}{11em}{Frustration - Adaptive Case} & Deciding next action & 2.75 & 1.59 & 2 \\
& & Understanding other agent's Intention & 2.95 & 1.57 & 2 \\
& & Adapting to other agent & 3 & 1.34 & 3.5 \\
& & Relying on the other agent & 3 & 1.52 & 3 \\ \cline{2-6}
& \multirow{4}{11em}{Frustration - Specified Orchestration} & Deciding next action & 2.2 & 1.4 & 1.5 \\
& & Understanding other agent's Intention & 2.6 & 1.35 & 2.5 \\
& & Adapting to other agent & 2.7 & 1.49 & 3 \\
& & Relying on the other agent & 2.6 & 1.6 & 2 \\ \hline
\end{tabular}%
}
% Descriptive
%\end{adjustbox}
\label{fig:subj-result}
\end{table*}

\clearpage
\newpage

\section{Experiment Details}
\label{app:experimentalsetup}

\subsection{Evaluation Domains}
\label{app:domains}

\subsubsection{\textbf{Multi-Agent Highway Orchestration Task:}} We first design a set of three multi-agent environments built on top of the Highway environment \cite{highway-env}. As shown in Figure \ref{fig:ma_highway_domain}, we construct a three lane highway where the human car (in orange) is on the top-most lane, and the AI agent (in purple) has to overtake and merge back into the human car's lane by increasing its speed and avoiding collisions with any other car in the scene. The human prefers that the AI agent car overtakes from the right-most (bottom-most in the figure) lane and not from the middle lane. Similarly, we construct two more scenarios with both cars in the right and middle lanes. In Figure \ref{fig:results_image}, these are referred to as MA Highway - Right Overtake, MA Highway - Left Overtake, and MA Highway - Middle Overtake, respectively.

\subsubsection{\textbf{Multi-Agent Locomotion Orchestration Tasks:}} Following prior works in constructing multi-agent environments \cite{peng2021facmac} in continuous control domains, we modify the MuJoCo-based locomotion environments \cite{todorov2012mujoco} for the human-AI teaming tasks which necessitate cooperation. Single-agent variant of these environments have been used as benchmark domains by single-agent PbRL works \cite{surf,rune,pebble}, further motivating their use. We construct multi-agent settings for Cheetah, Ant, Walker, Swimmer \& Hopper environments, as shown in Figure \ref{fig:ma_mujoco_domain}. In each environment, the human agent controls the \textcolor{orange}{orange highlighted joints} action space, while the rest of the joint angle controls are for the AI agent \textcolor{purple}{highlighted in purple}. For example, in the MA-Walker, the human controls the walker's right leg joints to walk forward, while the AI agent controls the left leg joints.

\subsection{State Space and Action Space}

\begin{table}[ht]
	\caption{MA MuJoCO - Action Space}
	\label{table:joint_angles}
    \resizebox{\columnwidth}{!}{
	\begin{tabular}{rll}\toprule
		\textbf{Environment} & \textbf{Human's Action Space} & \textbf{AI Agent's Action Space} \\ \midrule
		Walker & Right Leg Joints & Left Leg Joints \\
		Hopper & Thigh Rotor & Leg Rotor, Foot Rotor \\
		HalfCheetah & Front Rotor & Back Rotor \\
% Ant - Each Leg contains 2 joints %
		Ant & Front Left Leg, Right Back Leg & Front Right Leg, Back Leg \\
		Swimmer & First Rotor & Second Rotor \\ \bottomrule
	\end{tabular}
}  
\end{table}

We use the available MuJoCo implementation packaged by OpenAI Gym \cite{openai} to instantiate our environments. We use the default state space representation given by the package, and the same observation space is shared by both, the human and the AI agents. The details about the action space are given in Table \ref{table:joint_angles}.

For MA Highway, we use the Highway environment packages provided by \cite{highway-env} to create our custom multi-agent environments and the associated tasks. We use the continuous action space consisting of (steering, throttle) for both the agent cars.

\subsection{Reward Architecture and Embedding Space}
Following the implementation of \cite{pebble,rune,surf} we implement reward model via a neural network and bound the final output using a $\tanh$ activation function : [-1, 1]. For all the experiments, the reward model has three hidden layers with 256 neurons each followed by the prediction layer (with one neuron). We use the ADAM optimizer for training the SAC actor-critic as well as the reward model. The hyperparameters used for PEBBLE, RUNE \& SURF are given in Tables \ref{tab:pebble_hyperparams}, \ref{tab:surf_hyperparams}, and \ref{tab:rune_hyperparams}.

The input to the reward model is (state, action) tuple. As used previously \cite{surf}, we stack the state and action vectors and treat them as a single input for the reward model. 

\subsection{Human Policy \& Preference Oracles}
We assume an oracle scripted human who uses the underlying reward to correctly provide the binary feedback for their preferences. The feedback is given as follows : 
\begin{equation}
\label{eq:oraclelabel}
    y(\tau_0, \tau_1) = \begin{cases}
    0, & \text{if} \quad \sum_{i} \Tilde{R_h}(\tau_0) > \sum_{i} \Tilde{R_h}(\tau_1) \\ 
    1, & \text{if} \quad \sum_{i} \Tilde{R_h}(\tau_0) < \sum_{i} \Tilde{R_h}(\tau_1) \\ 
    \end{cases}
\end{equation}
where, $\Tilde{R_h}$ is the environment reward being used as the human preference reward function and $\tau_0, \tau_1$ are the queried trajectory pairs. Note that we work under the setting that the preference feedback is binary and therefore if the trajectory returns are equal we uniformly pick a preferred trajectory. This does not pose any problems with our chosen benchmark domains as the underlying reward is dense and shaped \cite{devlin2012dynamic,ng1999policy}.

To construct the human policy oracles, we train a SAC agent treating each environment as a single-agent environment, and retrieve the oracle policy for the human's action space from this SAC agent at the time of training the AI agent.

The hand-crafted rewards for MA Highway - Right Overtake have been shown in Table \ref{tab:highway_right_rewards}. Similarly, we construct the rewards for MA Highway - Middle \& Left Overtake domains.

\begin{table}[ht]
\centering
\caption{MA Highway - Right Overtake rewards. (x1,y1) is the position of AI agent and (x2,y2) is the position of the Human agent in the environment.}
\label{tab:highway_right_rewards}
\begin{tabular}{ccc}
\hline
\textbf{Condition}                                  & \textbf{Position}          & \textbf{Return} \\ \hline
Same Lane                                           & x1 \textless x2            & -0.5            \\
y1 \textgreater{}= y2 - 2 \& y1 \textless{}= y2 + 2 & x1 \textgreater x2         & 1               \\
Adjacent Lane                                       & x1 \textless{}= x2 - 10    & -0.25           \\
y1 \textgreater{}= 2\& y2 \textless{}= 6            & x1 \textgreater{}= x2 + 10 & 0.5             \\
                                                    & else                       & -1              \\
Overtaking Lane                                     & x1 \textless{}= x2         & 0.25            \\
y1 \textgreater{}= 6 \& y2 \textless{}= 10:         & x1 \textgreater{}= x2      & 0.25            \\ \hline
\end{tabular}
\end{table}

\subsection{Implementation, Code and Compute}
We use the publicly available implementation of \cite{pebble,rune,surf} for the implementation of SAC, PEBBLE, RUNE \& SURF algorithms. Our implementation of extending these PbRL algorithms to the multi-agent setting can be found in the code given in the supplementary. All the experiments were run on an Intel(R) Xeon(R) Gold 6258R CPU @ 2.70GHz, with Quadro RTX 8000 GPU.

\clearpage
\newpage

\section{Hyperparameters}
\label{app:hyperparams}

\begin{table}[ht]
\caption{PEBBLE Hyperparameters}
\label{tab:pebble_hyperparams}
\resizebox{\columnwidth}{!}{
\begin{tabular}{ll|ll}
\hline
\textbf{Hyperparameter} & \textbf{Value} & \textbf{Hyperparameter} & \textbf{Value} \\ \hline
Initial temperature            & 0.1               & Hidden units per each layer & 1024                               \\
Segment length                 & 30 (MA Highway)   & \# of layers                & 2                                  \\
                               & 50 (MA MuJoCo)    & Optimizer                   & Adam  \\
Learning rate                  & 0.0003            & Batch Size                  & 1024(DMControl)                    \\
Critic target update freq      & 2                 & Critic EMA $\tau$           & 0.005                              \\
$\beta_1, \beta_2$             & (0.9, 0.999)      & Discount $\gamma$           & 0.99                               \\
Frequency of feedback          & 500 (MA Highway)  & Maximum budget /            & 800/25 (MA Highway)                \\
                               & 10000 (MA MuJoCo) & \# of queries per session   & 1400/14 (MA MuJoCo)                \\
                               &                   & \# of pre-training steps    & 1000 (MA Highway)                  \\
\# of ensemble models $N_{en}$ & 3                 &                             & 5000 (MA MuJoCo)                   \\ \hline
\end{tabular}
}
\end{table}

\begin{table}[ht]
\centering
\caption{SURF Hyperparameters}
\label{tab:surf_hyperparams}
\begin{tabular}{ll}
\hline
\textbf{Hyperparameter}              & \textbf{Value}           \\ \hline
Unlabeled batch ratio $\mu$ & 4               \\
Threshold $\tau$            & 0.95            \\
Segment length              & 15 (MA Highway) \\
                            & 50 (MA MuJoCo)  \\
Loss weight $\lambda$       & 1               \\ \hline
\end{tabular}
\end{table}

\begin{table}[ht]
\centering
\caption{RUNE Hyperparameters}
\label{tab:rune_hyperparams}
\begin{tabular}{ll}
\hline
\textbf{Hyperparameter} & \textbf{Value}           \\ \hline
Segment length & 20 (MA Highway) \\
               & 50 (MA MuJoCo)  \\ \hline
\end{tabular}
\end{table}

\section{Additional Results}
\label{app:sec_additional_results}

\begin{figure*}[b]
\begin{center}
\centerline{\includegraphics[width=\textwidth]{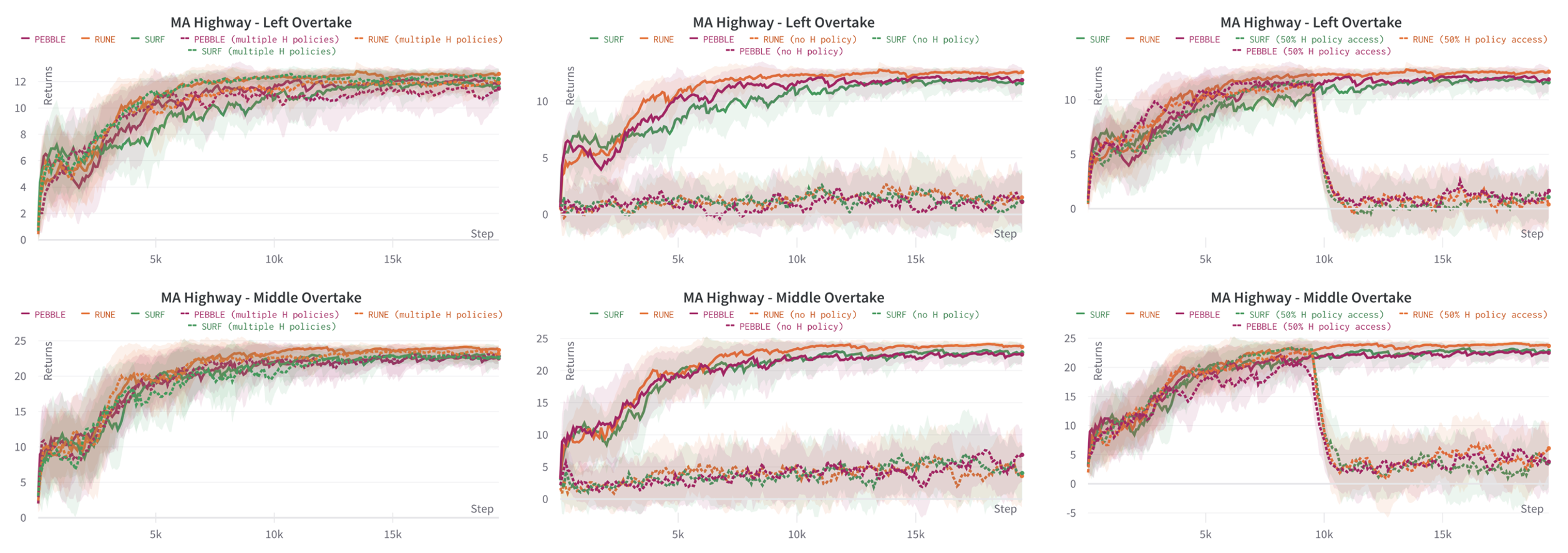}}
\caption{Learning curves on MA Highway-Left (\textit{row 1}) and MA Highway-Middle (\textit{row 2}): comparing (\textit{from L to R}) (a) Human Flexibility on multiple $\pi_H$, with Specified Orchestration case that assumes a single $\pi_H$ and complete access to it, (b) agent's 0\% access to $\pi_H$, and (c) partial access to $\pi_H$; as measured on the episodic returns. The solid lines and shaded regions represent the mean and standard deviation, respectively, across three runs.}
\label{fig:results_image_highway}
\end{center}
\end{figure*}

The learning curves on MA Highway-Left and MA Highway-Middle have been shown in Figure \ref{fig:results_image_highway}.

\clearpage
\newpage

\section{Societal Impact}
\label{sec:societal_impact}

This paper focuses on motivating the use of Preference-based Reinforcement Learning (PbRL) for human-AI teaming settings. We argue on two critical aspects of such teaming settings in this work, namely - Human Flexibility and AI agent's Access to Human Policy. The empirical benchmarking includes experiments across a suite of domains with simulated human oracles. Since this work uses simulated human oracle policies and preferences for the experiments, it does not bear any direct ethical implications. However, we note that once such studies advance to real-world teaming problems, ethical considerations with respect to privacy, adherence to the intended preference, consent, transparency, etc. will come in place.

Specifically, with the use of PbRL methods in real-world settings, we highlight that the human partner may experience unexpected and unwanted scenarios while interacting with the AI agent. This may include dealing with irrational or erratic behavior of the AI agent. This may further lead to frustration and fatigue for the human which may be undesirable. Moreover, trust on the AI agent can also play a crucial role in the effectiveness of modeling a human-AI teaming problem. For example, if the human partner does not trust the AI agent for task completion or its ability to learn their preferences (due to any possible reason), the human may find it difficult to give feedback that can be useful for the AI agent's training. Hence, generalizing PbRL for real-world human-AI teaming settings may require design decisions that can encompass these aspects to benefit the human partner agent.

\clearpage
\newpage

\end{document}